%% file: main.tex
\documentclass{article}
\input{header.tex}
\usepackage{comment}
\usepackage{fontawesome5}
\usepackage{booktabs}
\usepackage{multirow}
\usepackage{hyperref}
\usepackage{xurl} 
\usepackage{xcolor}

\usepackage[most]{tcolorbox}
\definecolor{logbg}{RGB}{247,249,253}
\definecolor{logframe}{RGB}{195,210,235}
\definecolor{toolbg}{RGB}{220,235,255}
\definecolor{addedbg}{RGB}{255,235,210}
\definecolor{addedtext}{RGB}{180,90,0}
\definecolor{phasebg}{RGB}{235,240,250}
\newcommand{\tool}[1]{%
  \colorbox{toolbg}{\textsf{\scriptsize\bfseries\strut #1}}}
\newcommand{\addedtool}[1]{%
  \colorbox{addedbg}{\textsf{\scriptsize\bfseries\strut
    {\color{addedtext}+}\,#1}}}
\newcommand{\phase}[1]{%
  \par\smallskip\colorbox{phasebg}{\small\sffamily\bfseries\strut\;#1\;}%
  \nopagebreak\smallskip}
  
\usepackage{enumitem}
\setlist[itemize]{topsep=0pt,itemsep=1pt,parsep=0pt,partopsep=0pt}

\newcommand{\email}[1]{\href{mailto:#1}{\nolinkurl{#1}}}

\newcommand{\FinchLogoo}{%
  \textcolor[RGB]{220,57,69}{\textsc{F}}%
  \textcolor[RGB]{220,57,69}{\textsc{i}}%
  \textcolor[RGB]{125,81,53}{\textsc{n}}%
  \textcolor[RGB]{30,104,36}{\textsc{c}}%
  \textcolor[RGB]{30,104,36}{\textsc{h}}%
}

\title{\includegraphics[width=0.05\textwidth]{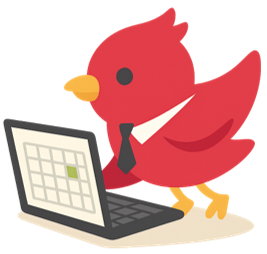}~ \FinchLogoo{}: Benchmarking Finance \& Accounting across Spreadsheet-Centric Enterprise Workflows}

\author{
\textsc{Finch} members\thanks{Correspondence to donghaoyu82@gmail.com; Full author list: Appendix~\ref{sec:author}.}
}

\begin{document}
\maketitle

\vspace{-1.5em}
\begin{center}
  \includegraphics[width=\linewidth]{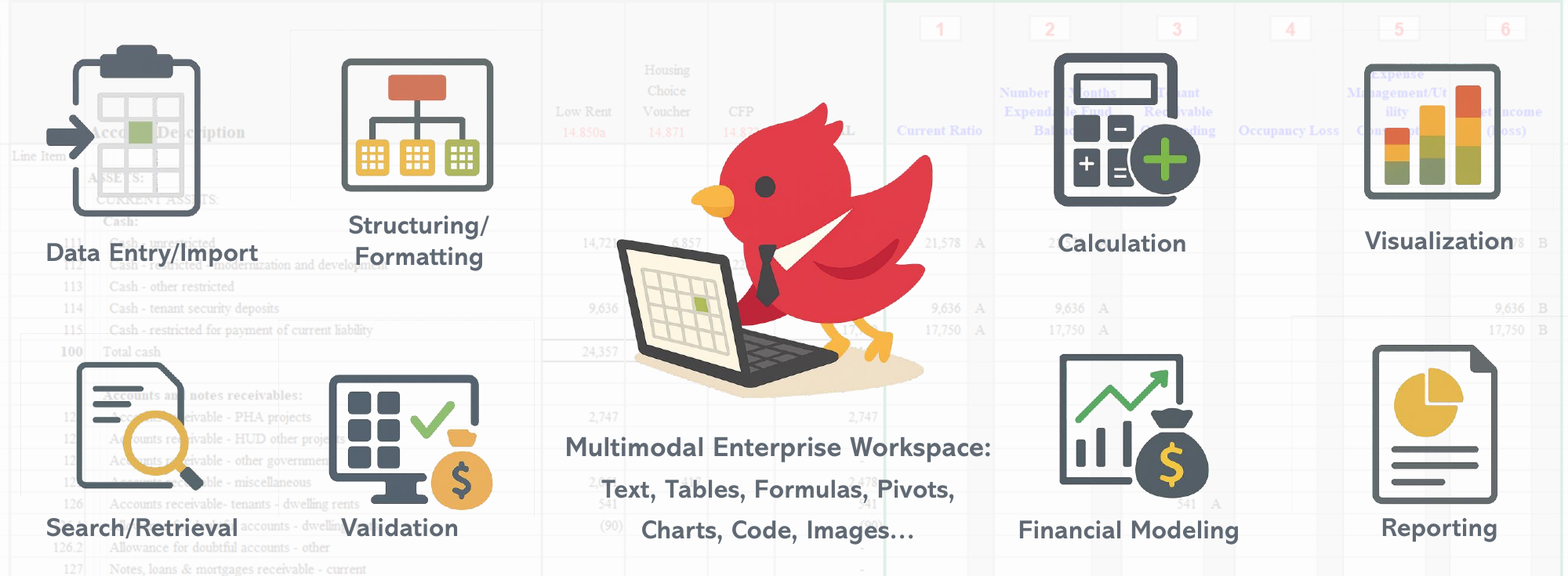}
  \captionof{figure}{Real-world F\&A work is messy, spanning heterogeneous and large-scale artifacts such as spreadsheets and PDFs. It's also long-horizon and knowledge-intensive: workflows interleave multiple tasks and span diverse domains such as budgeting, trading, asset management, and operations.}\label{fig:cover}
\end{center}

\vspace{0.5em}

\begin{abstract}
We introduce FinWorkBench (a.k.a. \textsc{Finch}) for evaluating agents on real-world, enterprise-grade finance and accounting workflows---interleaving data entry, structuring, formatting, web search, cross-file retrieval, calculation, modeling, validation, translation, visualization, and reporting. \textsc{Finch} is sourced from authentic enterprise workspaces from Enron (15,000 files and 500,000 emails) and across various financial institutions, covering the period 2000--2025, preserving the in-the-wild messiness of multimodal artifacts such as tables and charts, across diverse domains including budgeting, trading, and asset management.

We propose a workflow construction process that combines LLM-assisted mining of workflows from authentic enterprise environments with expert annotation: (1) LLM-assisted, expert-verified derivation of workflows from real-world email threads and version histories of spreadsheet files, and (2) meticulous annotation for workflows, requiring over 700 hours of expert effort. This yields 172 composite workflows with 384 tasks, involving 1,710 spreadsheets with 27 million cells, along with PDFs and other artifacts, capturing the intrinsically messy, long-horizon, knowledge-intensive, and collaborative nature of enterprise work.

We conduct both human and automated evaluations of frontier AI systems, including GPT~5.1, Claude Sonnet/Opus~4.5, Gemini~3~Pro, Grok~4, and Qwen~3~Max. GPT~5.1~Pro spends an average of 16.8 minutes per workflow yet passes only 38.4\% of workflows. Comprehensive case studies further surface the challenges that real-world enterprise workflows pose for AI agents.
\end{abstract}

\hypersetup{colorlinks=true, allcolors=black}

\vspace{-1.4em}
\begin{center}
\small

\href{https://huggingface.co/FinWorkBench}{%
  \raisebox{-0.2ex}{\includegraphics[height=1.1em]{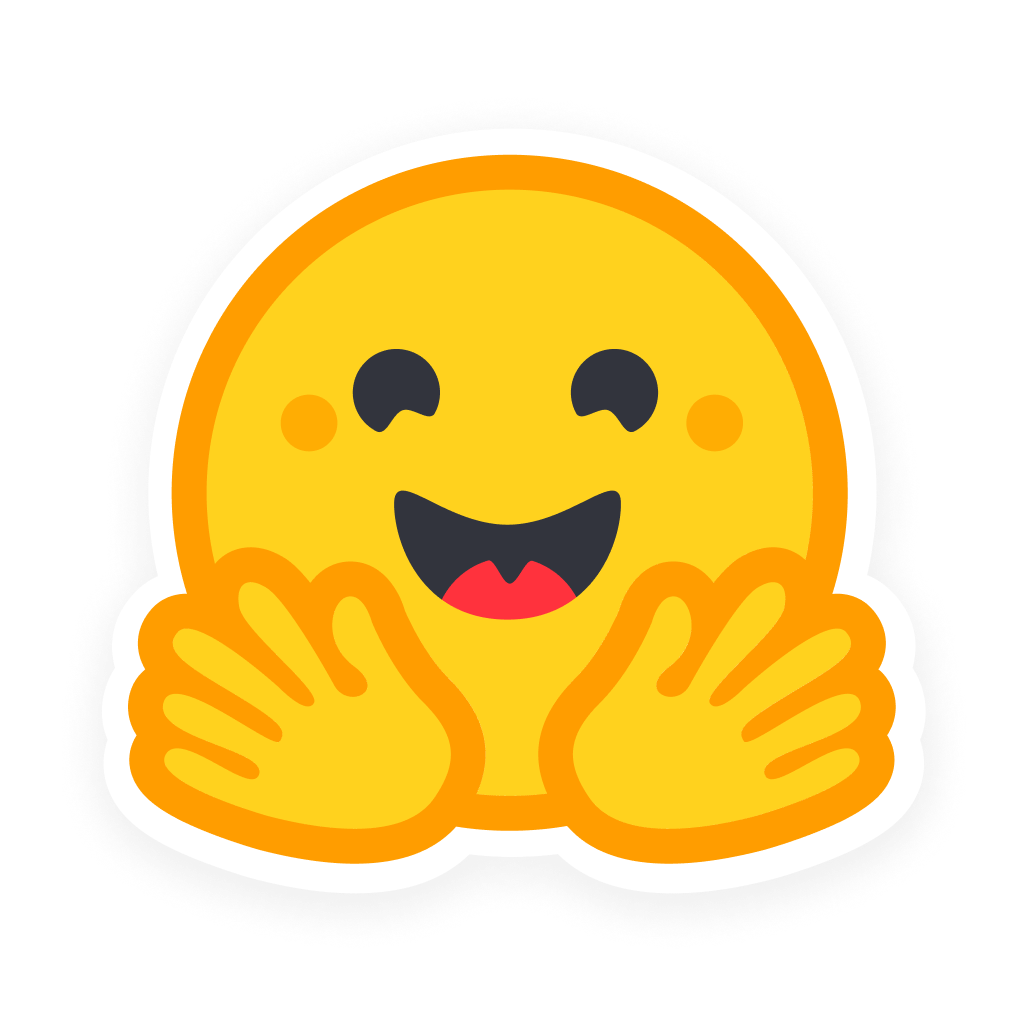}}\ \textbf{Dataset}: \url{https://huggingface.co/FinWorkBench}}\\

\end{center}

\clearpage 

\section{Introduction}

Frontier AI systems are increasingly transforming professional workspaces. AI-assisted tools like Claude~\citep{claude_cowork}, OpenClaw~\citep{openclaw}, ChatGPT~\citep{openai_agent},  Gemini~\citep{gemini_workspace}, and Copilot~\citep{microsoft_copilot} are now embedded in daily enterprise workflows---helping professionals draft documents, explore data, manipulate spreadsheets, and generate reports. These tools are particularly impactful in finance and accounting (F\&A), a high-stakes, knowledge- and labor-intensive domain critical to every organization.

However, real-world F\&A work is inherently \textbf{messy}, with substantial contextual complexity: artifacts are interconnected across heterogeneous spreadsheets, PDFs, and other artifacts, evolving through multiple versions with collaborative edits~\cite{klimt2004enron}; spreadsheets contain large, complex structures~\cite{spreadsheetllm2024} with cross-sheet references, intricate layouts, inconsistent formatting, cryptic terms, erroneous formulas, and multimodal artifacts such as charts, images, and code. It is also \textbf{long-horizon}~\cite{gdpval2025}: workflows demand multi-step reasoning spanning data entry, editing, retrieval, calculation, modeling, validation, reporting, and more.

This raises a key question: \textit{Can today's frontier AI agents actually handle the messy, long-horizon, and knowledge-intensive workflows that professionals face daily?}

To answer this, we introduce \textsc{Finch}, an enterprise-grade F\&A benchmark sourced from authentic enterprise environments. \textsc{Finch} captures the intrinsic complexity of professional work through:

\begin{itemize}
    \item \textbf{In-the-wild enterprise sourcing}: \textsc{Finch} is built around authentic enterprise spreadsheets, emails, and PDFs from real-world enterprise workspaces---primarily Enron~\cite{enrondata_pst} (about 15,000 spreadsheet files and 500,000 emails from 150 executives and employees) and EUSES~\cite{fisher2005euses} (about 450 financial spreadsheet files from various sources), along with financial institutions, global organizations, and government agencies, including the International Debt Report 2025, World Bank 2024 reports, Adidas 2024 financial statement, Public Accounts of Canada 2024~\cite{worldbank_idr_2024,financecanada_fiscalref_2025,hmt_pesa_2023}, etc. Documents are large, cross-referenced, and messy---containing rich multimodal artifacts such as tables, formulas, charts, pivots, and images.
    
    \item \textbf{Rigorous construction process}: We propose a novel workflow construction pipeline grounded in real collaborative context of emails and versioned artifacts. We induce workflows from enterprise email threads and attachments, where collaborators naturally describe, discuss, and track workflows as part of their daily work. We also propose an LLM-assisted, expert-verified method to derive workflows by analyzing changes across versioned spreadsheets, surfacing the underlying goals that drive professionals’ work. Annotators must reason over large multi-sheet workbooks, and subtle version deltas to infer what the original analyst was trying to achieve, making the annotation process substantially more difficult than curating QA pairs over isolated tables.
\end{itemize}

We compile 172 meticulously annotated, enterprise-grade workflows 
built on 1,710 spreadsheets, along with PDFs and other artifacts, 
collectively capturing the compositional, messy, knowledge-intensive, 
and collaborative nature of real work. Each workflow spans one or 
more interdependent tasks---data entry, editing, retrieval, 
calculation, modeling, validation, translation, visualization, 
and reporting---mirroring how professionals actually work on 
artifact manipulation and creation. Beyond their compositional task structure, \textsc{Finch} 
workflows demand substantial multi-step agent interaction: 
a case study on 20 representative workflows reveals that 
each requires a median of 16 tool calls (range: 6--107).

\begin{center}
  \includegraphics[width=\linewidth]{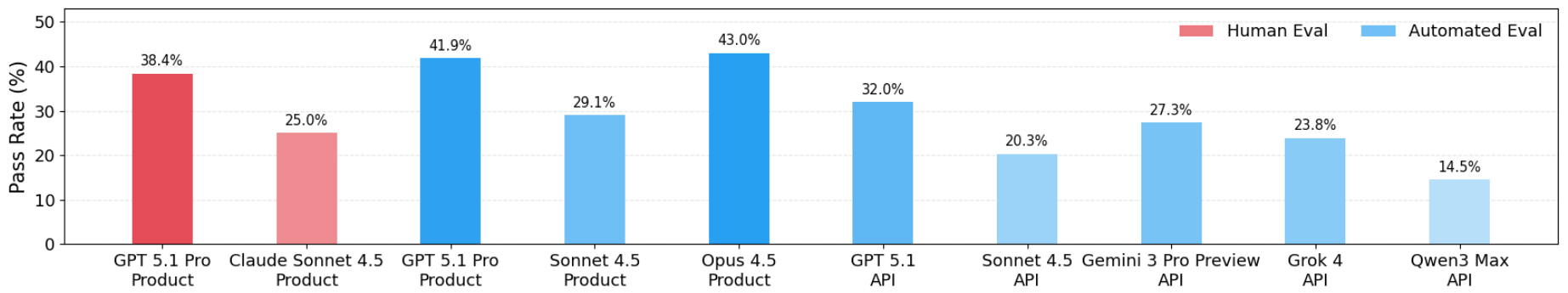}
  \captionof{figure}{Model pass‐rate comparison on \textsc{Finch} workflows. Bars show overall workflow success rates for product-side agents and API-based models. Detailed settings can be found in Section~\ref{sec:exp}.}\label{fig:intro_acc}
\end{center}

Evaluating such workflows poses nontrivial challenges, because \textsc{Finch} tasks usually involve complex and large spreadsheets and require assessing both structural correctness and semantic fidelity beyond exact matching. To address this, we provide both expert human evaluation and a scalable LLM-as-judge pipeline. Human evaluation serves as the gold standard, judging whether a workflow has been satisfactorily completed end-to-end. In parallel, we introduce an efficient multimodal LLM-as-judge evaluation method that compares inputs, model outputs, and reference artifacts using structured diffs, compact snapshots, and multimodal renderings, enabling reliable assessment at scale.

We evaluate a spectrum of frontier AI systems---including 
Claude Sonnet/Opus~4.5, GPT~5.1, Gemini~3, Grok~4, and 
Qwen~3---using both expert evaluation and our automated 
pipeline. Even the strongest frontier agents pass fewer 
than 50\% of \textsc{Finch} workflows: GPT~5.1~Pro spends 
an average of 16.8 minutes per workflow yet achieves only 
38.4\% under human evaluation.

Case analyses further reveal how each dimension of 
real-world complexity contributes to failure.
Long-horizon composition is a key bottleneck: 
GPT~5.1~Pro drops from 44.3\% on workflows with 
$\leq$2 tasks to 23.5\% on those with $>$2 tasks, 
as errors accumulate across steps.
The messy, irregular structure of enterprise 
spreadsheets leads to frequent data retrieval errors, and
agents often struggle to reconstruct latent 
knowledge-intensive business logic encoded in spreadsheet formulas.
Multimodal artifacts further amplify difficulty: 
on the 20 workflows involving PDFs, images, or Word 
documents, GPT~5.1~Pro's pass rate drops to 35.0\%, 
below its overall rate.
Taken together, these results suggest that it is the 
\textbf{combination} of messy inputs, long-horizon 
dependencies, domain knowledge, and heterogeneous 
formats---rather than any single factor---that drives 
the sharp performance degradation on real enterprise 
workflows.

\begin{center}
  \includegraphics[width=\linewidth]{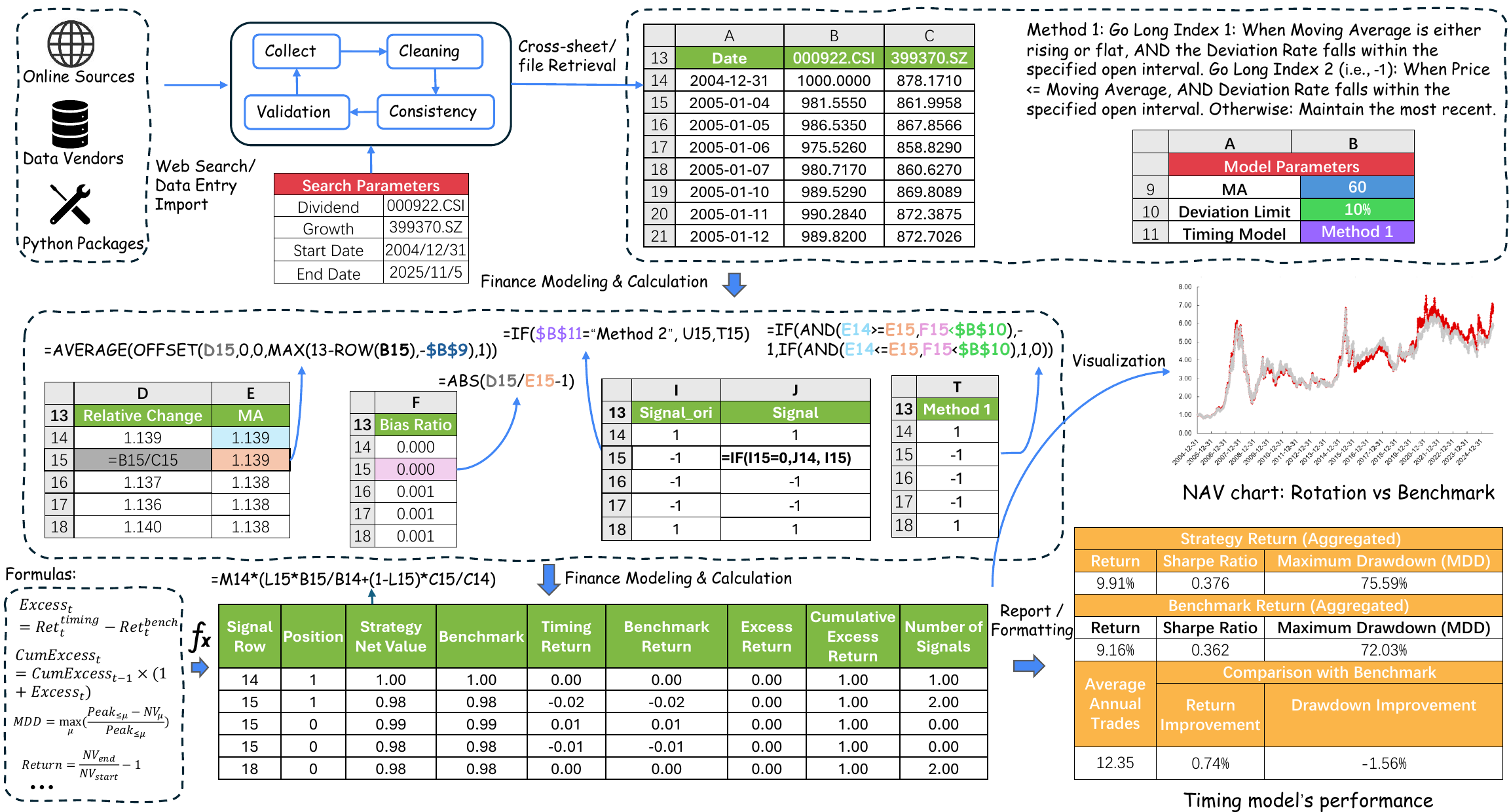}
  \captionof{figure}{Illustration of an end-to-end predictive modeling workflow typically performed by financial analysts. The workflow involves multiple steps, including web search, data import, cross-sheet and cross-file retrieval, calculation and financial modeling, visualization, and report generation. More illustrative examples for data characteristics in \textsc{Finch} are presented in Appendix~\ref{sec:example}.
}
  \label{fig:example}
\end{center}

\section{\FinchLogoo{}: A Real-world Finance \& Accounting Workflow Benchmark}
\subsection{Dataset Construction}

Echoing the existentialist dictum that existence precedes essence~\cite{sartre1946existentialism}, we argue that professional work should be observed in real-world enterprise environment before it is formally defined. Motivated by this perspective, we propose a novel workflow construction pipeline grounded in the authentic collaborative context of emails, versioned spreadsheets, and final deliverable spreadsheets and reports, as illustrated in Figure~\ref{fig:labeling} (a--c). First, we induce workflows from enterprise email threads and versioned documents, where collaborators naturally describe, discuss, and track work as part of their daily routines. Second, we derive workflows by analyzing changes across versioned spreadsheets, surfacing the actual data transformations and analysis steps that professionals performed. Third, we leverage high-quality spreadsheets and reports: we design workflows, author task instructions, and revise these spreadsheets and reports so that they serve as the input files and reference solutions.

\begin{center}
  \includegraphics[width=\linewidth]{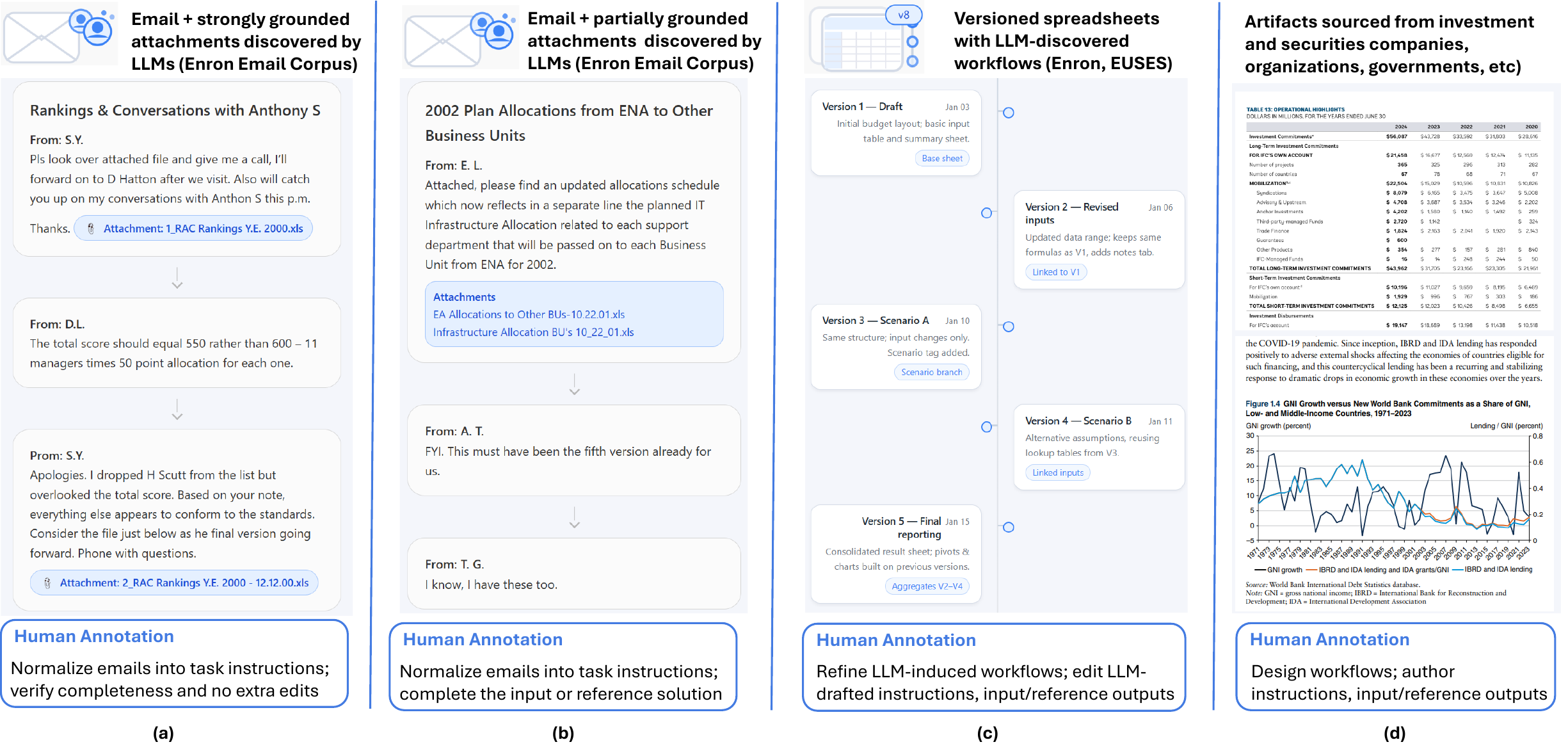}
  \captionof{figure}{Illustration of our workflow construction pipeline from real-world enterprise emails, versioned spreadsheets, and high-quality artifacts.}\label{fig:labeling}
\end{center}

All annotated workflows from these different channels are consolidated into a unified schema with consistent fields (NL instruction, input files, reference outputs), and each workflow is tagged with task types (e.g., data entry/import, structuring, validation) and business types (e.g., planning and budgeting, pricing and valuation, operations, asset management). Note that the reference outputs may include both file-based reference answers (for most generation/editing cases) and textual reference answers (for a small number of QA and summary/visualization cases).

\subsubsection{Workflow from Enterprise Email Threads}
We first mine real-world enterprise email threads to surface workflows.  
Starting from the Enron Email Corpus, we prompt GPT-5 to identify collaborative messages that (i) explicitly state a business goal (e.g., ``update the RAC rankings'' or ``revise the 2002 allocations'') and (ii) reference one or more attached spreadsheets.  
For each selected thread, the model summarizes the communicative intent and articulates a workflow description.

In the \emph{strongly grounded} case illustrated in Figure~\ref{fig:labeling} (a), both the input and the final reference artifacts for the workflow are already present as attachments in the thread (e.g., an initial ranking file and a corrected version). Strongly grounded cases were a primary motivation of this benchmark, but they are relatively rare.  
In the \emph{partially grounded} case illustrated in Figure~\ref{fig:labeling} (b), the email specifies a clear goal, but only some of the required artifacts are attached (e.g., just the updated schedule, or an intermediate report).  
Across both cases, human experts normalize conversational email text and LLM-drafted descriptions into workflow instructions and abstract away idiosyncratic details while preserving the business intent.
For strongly grounded threads, annotators primarily verify that the attached files exactly implement the requested change without extra edits.  
For partially grounded threads, they either identify the missing artifacts from attached spreadsheets and revise them to align with the described workflow---carefully avoiding the introduction of new changes---or create the missing artifacts themselves, which typically requires much more effort.

\subsubsection{Workflow Derivation from Versioned Spreadsheets}
Beyond explicit messages in email threads, we propose to discover workflows that are implicitly captured in spreadsheet version histories, as illustrated in Figure~\ref{fig:labeling}(c).
We collect families of versioned workbooks from the Enron and EUSES repositories and apply an LLM-based differencing procedure that recognizes consecutive versions and infers the underlying workflow.

For each recognized pair (or chain) of versions, we prompt GPT-5 to propose (i) one or more workflow types (e.g., ``date-stamped versioning, assumption updates, and error correction'', ``data entry, structuring, and visualization'') and (ii) a detailed NL description of all changes.  
Human experts then validate and refine these LLM-induced workflow candidates. They first determine whether the proposed diffs constitute a coherent and meaningful workflow rather than incidental churn. For accepted cases, they (i) rewrite the draft description into a precise task instruction that describes the transformation, and (ii) edit the corresponding workbook versions so that the designated input and reference files cleanly realize the described workflow without introducing out-of-scope changes beyond the instruction.  
The corresponding input and reference files are thus grounded in the actual versions used in the diff, yielding workflow instances that do not rely on email context but are anchored in real enterprise spreadsheet evolution.

\subsubsection{Workflow Sourced from Final Deliverable Spreadsheets and Reports}

Finally, we curate workflows from high-quality spreadsheets and reports drawn from the Enron and EUSES corpora, various investment and securities companies, international organizations, and national governments (e.g., the World Bank and the Canadian and British governments). Domain experts author realistic workflow instructions and construct input and reference files based on final deliverable artifacts. For example, a valuation model from an investment firm can be turned into a financial modeling task; a World Bank report can be used to define a data-driven summarization and visualization task; and bilingual reports from the Canadian government can be used to construct translation and consistency-checking tasks. We additionally leverage labeled samples from existing datasets: we adapt 10 financial cases from WideSearch~\cite{wong2025widesearch} into web-search-centric workflows and extend them into multi-step calculation and visualization pipelines, and we leverage 3 examples from DABStep~\cite{egg2025dabstep} to construct multi-source question answering workflows. We rely on WideSearch’s annotations without further revision, as they are supported by extensive web-crawled information available for validation.

\subsubsection{Quality Control}
Given the high complexity of each workflow, we adopt a rigorous multi-stage quality control process. All workflows are annotated and reviewed by a team of five experts: two annotators with interdisciplinary backgrounds in finance and computer science, and three annotators with computer science backgrounds. Among them, three annotators (including one female) have over nine years of industry experience, and two are outstanding Ph.D. or master’s students.

Annotators are instructed to skip workflows that are similar to existing ones to maximize diversity. 20\% of our annotators are full-time reviewers who do not create any original annotations, and 80\% act as full-time workflow creators (producing the initial annotations) and also serve as part-time reviewers. Each workflow is proposed by one person and typically reviewed by two others, including a dedicated full-time reviewer who reviews all queries and also inspects LLM-generated review comments to surface potential annotation issues. Importantly, Quality Control (QC) is iterative rather than one-pass: approximately 40\% of workflows undergo at least one round of revision; 20+ workflows went through three or more QC rounds, and we held regular group discussions to resolve challenging cases and align standards. In addition, ChatGPT 5.1 Pro and Claude Sonnet 4.5 are used as secondary checkers to flag potential issues, which are always verified by human experts.
Together, these procedures required over 700 hours of expert annotation.

\subsection{Dataset Characteristics }
\label{sec:data_ana}

\textsc{Finch} comprises 172 meticulously annotated, enterprise-grade workflows that collectively capture the compositional, messy, multimodal, and collaborative nature of real finance and accounting work. Across these workflows, the corpus contains 1{,}710 spreadsheets (956 distinct ones in 301 Excel files) together with 17 PDFs, 12 images, 3 Word documents, as well as JSON, CSV, and Markdown. This mixture reflects how real analysts coordinate over heterogeneous artifacts rather than clean, single-table inputs.

\begin{center}
\includegraphics[width=\linewidth]{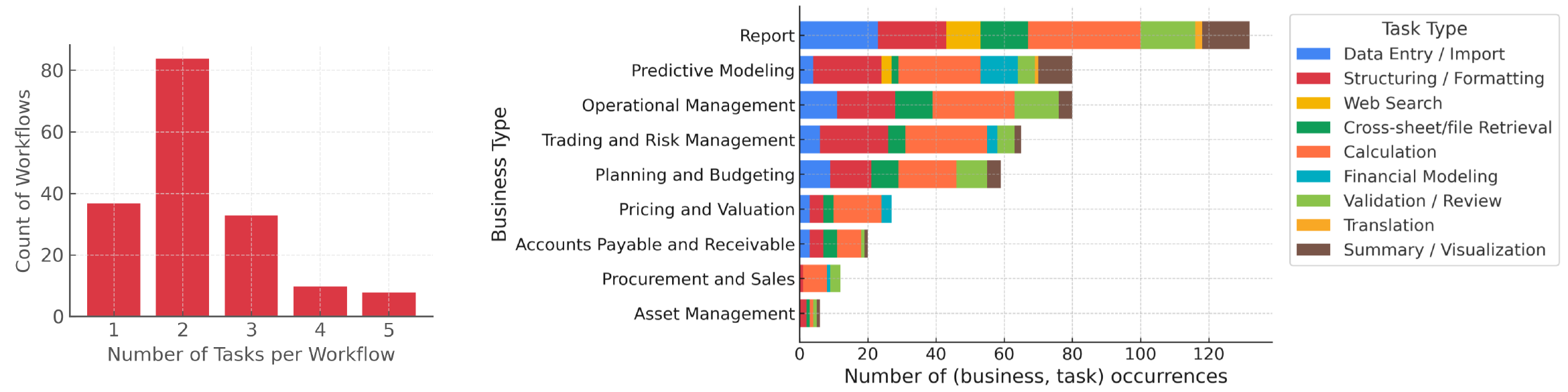}
  \captionof{figure}{Distribution of number of tasks per workflow and task types across business types.}\label{fig:biz_distribution}
\end{center}

Figure~\ref{fig:biz_distribution} summarizes the coverage of task and business types. On the task side, categories are:
\begin{itemize}
  \item Calculation (119 workflows): filling in formulas or computing figures (e.g., net value).
  \item Structuring / Formatting (86): reorganizing tables (e.g., adjusting hierarchies), formatting content (e.g., font size and cell fill), and inserting/deleting rows or columns.
  \item Data Entry / Import (44): transcribing or importing data from spreadsheets, PDFs, images, or external sources into spreadsheets.
  \item Validation / Review (37): checking consistency and reconciling calculations within a sheet or across sheets/files.
  \item Cross-sheet/file Retrieval (36): pulling values from multiple sheets or files into a target workbook.
  \item Summary / Visualization (33): producing summaries or charts that surface key financial insights.
  \item Financial Modeling (15): extending or calibrating valuation and timing models, often via scenario and sensitivity analysis.
  \item Web Search (11): collecting financial data from the web and integrating it into spreadsheets.
  \item Translation (3): translating spreadsheets or reports while preserving structure, formatting, and layout.
\end{itemize}

On the business side, workflows span reporting (48 workflows), trading and risk management (35), predictive modeling (33), operational management (36), planning and budgeting (26), pricing and valuation (15), accounts payable/receivable (10), as well as procurement and sales (7) and asset management (3); some workflows are tagged with multiple business types. Overall, the distribution indicates that \textsc{Finch} targets core finance and accounting verticals rather than curated toy tasks.

\subsubsection{Task Compositionality}

\textsc{Finch} is explicitly designed around composite workflows rather than isolated tasks. As shown in Figure~\ref{fig:biz_distribution}, only 37 workflows (21.5\%) are single-task; the remaining 135 (78.5\%) involve multiple tasks. These tasks are not independent subtasks, but are interleaved around shared spreadsheets and files. A typical workflow may begin with structuring or importing raw data, proceed to cross-sheet or cross-file retrieval, and then culminate in calculations, modeling, or reporting. The distribution in Figure~\ref{fig:biz_distribution} shows that most workflows weave multiple tasks.

Importantly, each “task” itself typically requires substantial multi-turn reasoning: for example, web search often entails many rounds of LLM calls to discover, filter, and verify evidence; cross-sheet retrieval requires iterative calls to read and locate key information across multiple sheets; and calculation usually spans many formulas distributed over different rows and columns.

To further characterize the practical complexity of \textsc{Finch} workflows, 
we conduct a case study on tool-call overhead using Claude Coworker (Opus~4.6) 
on 20 representative tasks (Appendix~\ref{appendix:tool_call}). Excluding two extremely tool-intensive web-search workflows that require 71 and 107 tool calls respectively, the 
remaining tasks range from \textbf{6 to 25 tool calls} (median~14). 
Notably, tool-call overhead does not scale linearly with the number of tasks; intrinsic task complexity---such as the depth of 
business-logic reasoning and cross-sheet dependencies---is a stronger driver. 
These findings suggest that even single- or two-task workflows can demand 
substantial multi-step agent interaction, reinforcing that \textsc{Finch} 
captures meaningful operational complexity beyond what task counts alone convey.

\subsubsection{Messiness}

The source files in \textsc{Finch} are intentionally large, multi-sheet, and structurally complex. At the file level, 86.6\% of workflows involve more than one file when counting both input and reference artifacts, and a workflow touches up to 14 distinct files. At the spreadsheet level, 92.4\% of workflows involve multiple input and reference sheets, with an average of 8 sheets and a long tail reaching 91 sheets. As a result, systems must navigate cross-sheet dependencies, hidden logic, and scattered intermediate calculations rather than operating on a single “analysis” sheet. Moreover, most spreadsheets exhibit complex layouts that interleave text, numerical values, formulas, and charts, as well as intricate single- or multi- table structures with nested headers, hierarchical data, merged cells, blank rows and columns, and other irregularities.

Cell-level statistics further highlight the scale of the data. The median workflow covers 15K cells (157K on average for all workflows), with the largest one scaling to 3.7 million cells. Formula density is similarly skewed: while workflows contain an average of 21.5K formulas (the median is 212), reflecting deeply nested calculations and long dependency chains. Taken together, these properties create a challenging regime in which models must reason over large, noisy, and highly irregular spreadsheet layouts, rather than clean, rectangular tables.

\begin{center}
  \includegraphics[width=\linewidth]{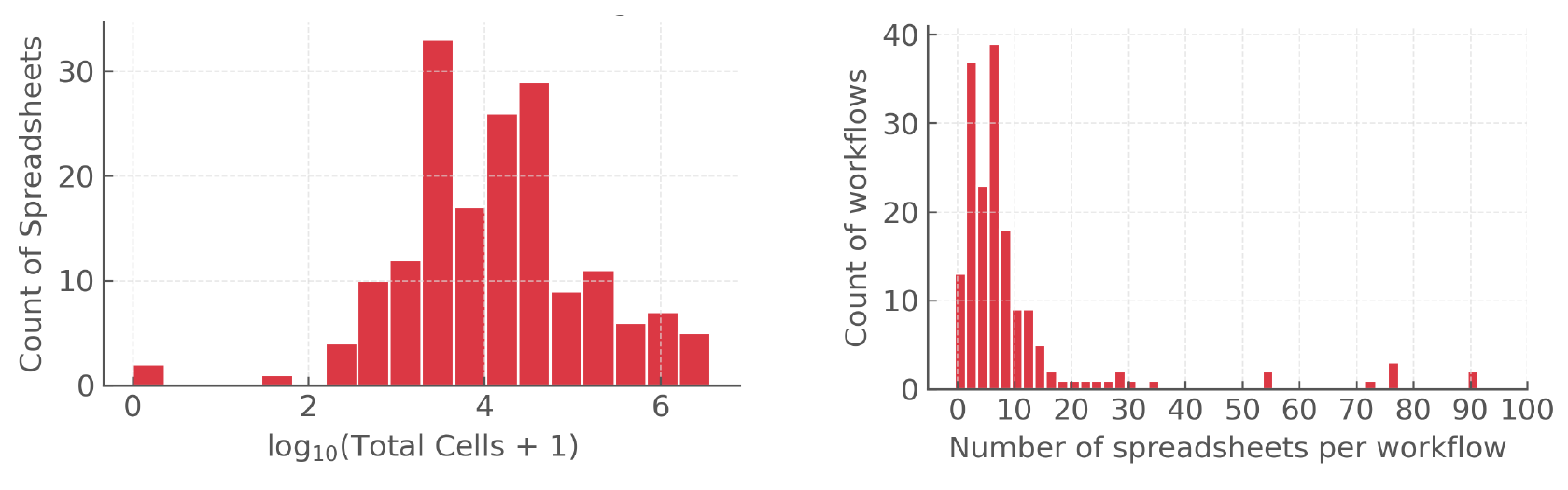}
  \captionof{figure}{Distribution of the number of sheets and cells per workflow.}\label{fig:moredistribution}
\end{center}

\subsubsection{Multimodality}

Although \textsc{Finch} is spreadsheet-centric, the workflows are inherently multimodal. Around 10.5\% of workflows link spreadsheets with additional non-spreadsheet artifacts such as PDFs, Word documents, and images, and 7.6\% explicitly require reasoning over PDFs or images. Within the spreadsheets themselves, 20.3\% of workflows include charts and 2.3\% feature pivot tables, so models must understand not only raw cell values but also derived visual summaries and explicit aggregation structures (and most workflows involve implicit aggregation structures).
This multimodal, cross-artifact structure stands in contrast to prior benchmarks that operate purely on isolated tables, and better reflects the environments in which enterprise finance and accounting tasks actually occur.

\subsection{Evaluation Method}

\subsubsection{Human Evaluation}
We conduct human evaluation on all workflows to directly assess model performance. For each workflow, annotators read the NL instruction, inspect the input, reference, and model output files side by side (typically by aligning spreadsheets or documents in adjacent tabs), and determine whether the model has faithfully completed the requested job. A workflow is marked as successful only if the model generates or revises the content and structure in accordance with the instruction and no critical errors, omissions, or unintended changes are introduced; otherwise, it is labeled as a failure. Importantly, evaluation is based on whether the instruction has been satisfactorily fulfilled rather than on a purely mechanical comparison between model and reference outputs, since there may be multiple acceptable solutions for summarization, visualization, formatting, formulas, and related aspects. To reduce subjectivity and ambiguity, annotators ultimately assign a binary pass/fail label for each workflow. These human judgments serve as the gold standard for measuring model performance and for validating the reliability of our automatic evaluation method.

\begin{center}
  \includegraphics[width=\linewidth]{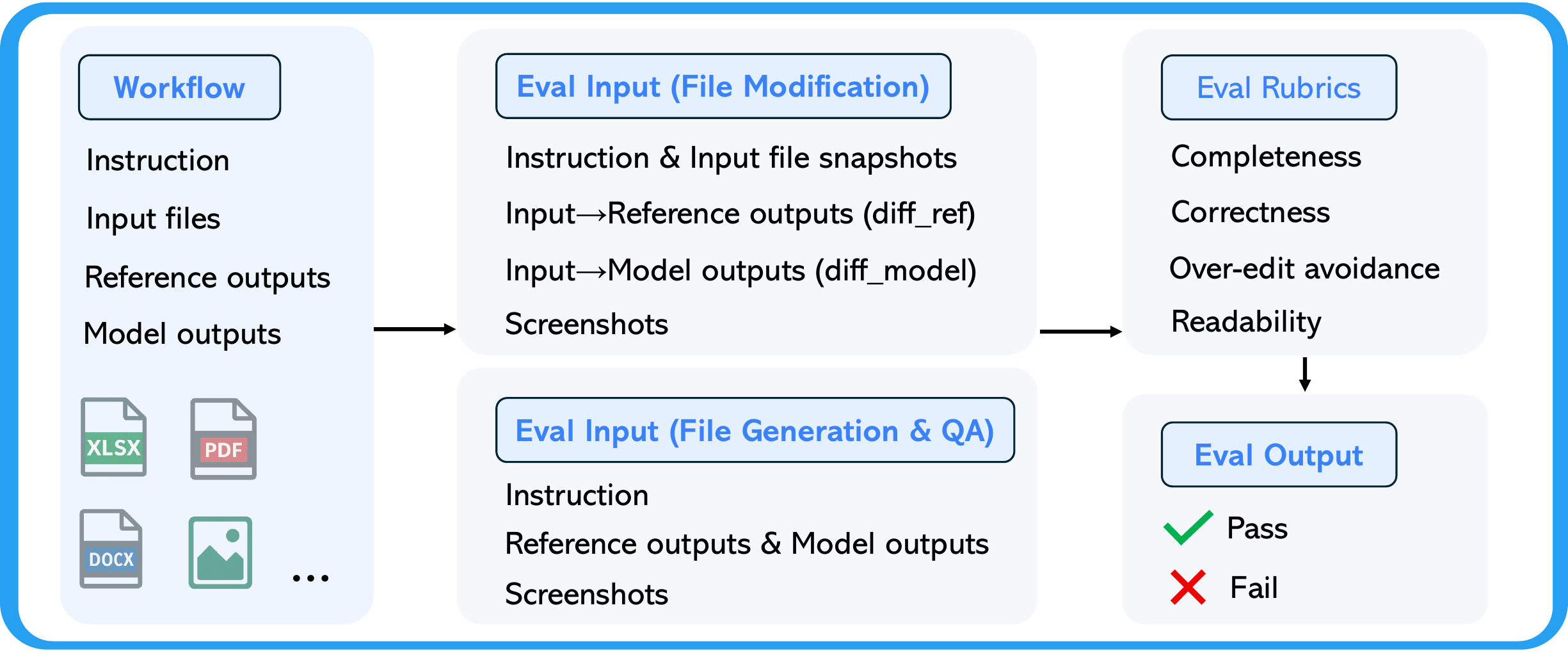}
  \captionof{figure}{Illustration of our automated evaluation pipeline. Here, \texttt{diff\_ref} denotes the diff between the input file and the reference output, and \texttt{diff\_model} denotes the diff between the input and the model output. We categorize all workflows into file modification, file generation, and file QA. This categorization is independent of the task types in Section~\ref{sec:data_ana}; for example, a \emph{calculation} task may generate a new file, modify an existing one, or simply return a textual answer.}\label{fig:evaluation}
\end{center}

\subsubsection{LLM-as-Judge Evaluation}
To scale evaluation, we employ an LLM-as-judge framework that supports the three high-level task types in \textsc{Finch}: \emph{modify} (editing input artifacts), \emph{generate} (creating new files such as workbooks and documents), and \emph{QA} (answering questions based on one or more artifacts). The framework accepts heterogeneous inputs---including \texttt{.xlsx}, \texttt{.txt}, \texttt{.docx}, \texttt{.md}, \texttt{.pdf}, and images---and normalizes them into a sequence of textual inputs and screenshot images for the judge model.

For modification tasks, especially on spreadsheets, the framework computes structured diffs between the input and the reference output (\texttt{diff\_ref}) and between the input and the model output (\texttt{diff\_model}), and then builds a compact “input snapshot” (\texttt{snapshot}) that, for each modified sheet, retains only the first and last ten rows and the first five columns (which typically capture table headers and layout) together with rows and columns that contain edited cells. This preserves the crucial context for \texttt{diff\_ref} and \texttt{diff\_model} while dramatically reducing token length. In parallel, the framework renders screenshots (\texttt{screenshot}) of sheets containing changes from the input, reference, and model output, so that the judge perceives merged cells, conditional formatting, charts, and other layout-sensitive properties that are difficult to encode as text alone.

For generation tasks involving spreadsheets, the framework extracts all cell values and formulas from both the reference and the model output, and captures screenshots of every sheet, since the entire generated artifact must be verified rather than just localized edits. For QA tasks, it feeds the reference answer and the model’s response, optionally augmented with relevant input artifacts when the question requires grounding in input artifacts.

We design three task-specific judge prompts for modify, generate, and QA, respectively, but they share a common evaluation rubric. In all cases, the judge is instructed to focus on (i) \emph{completeness} with respect to the NL instruction, (ii) \emph{numerical and logical correctness} of derived values and formulas, (iii) the \emph{over-edit avoidance}, penalizing unnecessary and unexpected changes of the workbook beyond instruction, and (iv) readability of the formatting and structure. Exact cell-by-cell equality with the reference is not required when multiple solutions are acceptable (e.g., alternative layouts, equivalent formulas, or different but semantically equivalent summaries); instead, the judge decides whether the model has satisfactorily fulfilled the instruction. To reduce subjectivity, the judge outputs a binary score (pass/fail) along with a short NL rationale. In some web search tasks, the rubric permits small tolerance bands when comparing values, allowing for discrepancies in data from different sources.

This LLM-as-judge framework not only automates large-scale evaluation but also surfaces subtle spreadsheet errors (such as formulas silently replaced with static values) that are difficult to catch with GUI-based human inspection alone. In Section~\ref{sec:exp_result}, we report the consistency between human and automated evaluations and show that the LLM-as-judge scores closely track human judgments. Looking forward, \textbf{agentic evaluation} is quite promising for future work for flexible workflow evaluation.

\section{Experiments}
\label{sec:exp}
\subsection{Agents and Models}

\subsubsection{Product-side Agents}

We evaluate two frontier product-side agents: (i) \emph{ChatGPT} using the GPT 5.1 model in Pro mode, and (ii) \emph{Claude} using the Sonnet/Opus 4.5 models in thinking mode. We focus on these two systems rather than alternatives such as Gemini or Grok because they natively support returning downloadable files (e.g., spreadsheets) as outputs, rather than emitting code or markdown-formatted tables that are not intuitive for human evaluation. For both agents, we enable their external web browsing, but disable using historical chats so that each workflow is evaluated independently and without cross-run leakage. Since model updates are frequent and manual evaluation is very time-consuming, we used the latest model from our final round of experiments and did not consider subsequent updates.

\subsubsection{API-based Models}

We evaluate five frontier LLMs via API interfaces (Table~\ref{tab:model_config}). We adopt SpreadsheetBench~\citep{ma2024spreadsheetbench} as the baseline framework because it provides a principled code-generation paradigm for spreadsheet-centric tasks, treating executable code as the model’s action space and enabling direct manipulation of spreadsheets through standard libraries. This design naturally aligns with \textsc{Finch}, where workflows require complex spreadsheet operations, formula reasoning, and cross-sheet dependencies that cannot be reliably handled by text-only outputs.

While SpreadsheetBench was originally designed for relatively small and clean spreadsheets, we extend it with richer spreadsheet encodings, multimodal input support, and stricter execution and evaluation protocols, allowing it to scale to the large, messy, and long-horizon enterprise workflows in \textsc{Finch}.

\begin{table}[h]
\centering
\small
\begin{tabular}{lccccc}
\toprule
\textbf{Model} & \textbf{Provider} & \textbf{Context} & \textbf{Max Output} & \textbf{Vision} & \textbf{Native PDF} \\
\midrule
GPT 5.1~\cite{gpt5openai} & OpenAI & 400K & 128K & \checkmark & \checkmark \\
Claude Sonnet 4.5~\cite{claudesonnet45} & Anthropic & 1M$^\dagger$ & 64K & \checkmark & \checkmark \\
Grok 4~\cite{grok4} & xAI & 256K & 256K & \checkmark & --- \\
Qwen 3 Max~\cite{qwen3technicalreport} & Alibaba & 256K & 32.8K & --- & --- \\
Gemini 3 Pro Preview~\cite{gemini3pro} & Google & 1.05M & 65.5K & \checkmark & \checkmark \\
\bottomrule
\end{tabular}
\caption{API-based model configurations. Context and output limits are measured in tokens. Vision indicates native image input support, while Native PDF refers to direct PDF file ingestion via the provider's API without explicit text extraction. $^\dagger$Available via long-context beta API mode.}
\label{tab:model_config}
\end{table}
\paragraph{Spreadsheet Encoding.} SpreadsheetBench produces text tables without preserving cell addresses, data types, or formulas. However, these details are essential for tasks in \textsc{Finch}. We extend SpreadsheetLLM~\cite{spreadsheetllm2024} encoding and introduce a \emph{semantic-rich tuple encoding} that preserves full structural and semantic fidelity. Each sheet begins with its name and the corresponding data bounding box (e.g. \texttt{\#\# Sheet: [name] (A1:Z100)}). We then serialize the bounded region using a Markdown-based format. Each cell is encoded as a tuple \texttt{(Address, Value, Type, Formula)}, where \texttt{Address} denotes the cell reference (e.g. \texttt{A3}), \texttt{Type} indicates the data type (\texttt{T} = Text, \texttt{I} = Integer, \texttt{F} = Float, \texttt{D} = Date, \texttt{B} = Boolean), and \texttt{Formula} records the cell formula (e.g., \texttt{=SUM(A1:A10)->100}).


\paragraph{Multimodal Input Handling.}
We extend the framework to support multimodal inputs involving images and PDFs. For vision-capable models (GPT 5.1, Claude Sonnet 4.5, Grok 4, and Gemini 3 Pro), we use each provider's official multimodal API to transmit visual inputs alongside text prompts. For PDF documents, we adopt a tiered strategy. Models with native PDF support---GPT 5.1, Claude Sonnet 4.5, and Gemini 3 Pro Preview---directly ingest PDF files via their file upload interfaces, enabling analysis of both textual and visual elements without pre-extraction. For Grok 4, which lacks native PDF support, we extract text using PyMuPDF and include it in the \texttt{pdf\_content} field. For Qwen 3 Max, which lacks multimodal support entirely, both image and PDF content are converted to textual descriptions. While this fallback retains semantic cues, it loses layout and visual context.

\paragraph{Context Management.}
To handle large spreadsheets that may exceed model context limits, we implement automatic truncation. We reserve 32K tokens for model output—sufficient for comprehensive code generation and analysis while remaining within the output limits of all evaluated models. Truncation is triggered when input exceeds the remaining capacity, removing content from the end of spreadsheet data with an explicit notice appended to inform the model of data loss.



\subsection{Experimental Results}
\label{sec:exp_result}

\paragraph{Product-side agents (ChatGPT 5.1 Pro vs. Claude Opus/Sonnet 4.5).} As shown in Figure~\ref{fig:intro_acc} and Table~\ref{tab:eval_results}, ChatGPT~5.1~Pro and Claude Opus~4.5 achieve the strongest overall pass rates on \textsc{Finch}.
Their advantage largely comes from rich interactive affordances: they can iteratively inspect spreadsheets, revise intermediate states, and recover from partial errors over many tool calls. However, they solve fewer than 50\% of the \textsc{Finch} workflows, suggesting that real-world finance and accounting work remains far from ``solved'' even for frontier agents.
The detailed analysis in Figure~\ref{fig:task_breakdown} further highlights that long-horizon composition is a key bottleneck: when a workflow contains more than two tasks, the pass rate drops sharply---GPT 5.1 Pro decreases from 44.3\% (workflows with $\leq 2$ tasks) to 23.5\% (workflows with $>2$ tasks), and Claude Sonnet 4.5 decreases from 30.3\% to 11.8\%. This indicates that error accumulation across steps and missing intermediate affordances disproportionately hurt multi-step execution.

Pass rate also varies substantially by task type (Figure~\ref{fig:task_breakdown}).
Data Entry / Import and Structuring / Formatting are consistently among the most challenging categories, which aligns with \textsc{Finch} spreadsheets exhibiting messy layouts, irregular tables, and nontrivial structural constraints.
Moreover, Data Entry / Import workflows are frequently entangled with web search or PDF parsing, introducing multimodal dependencies that amplify failure modes.
Notably, Translation---a task where modern LLMs typically excel in standard NLP settings---performs surprisingly poorly in \textsc{Finch}.
In finance-heavy tables, translation can easily distort or drop critical structure and layout cues (e.g., header hierarchies, row/column alignment), and large grids make omissions more likely, leading to systematic failures.
Detailed error analysis can be found in Section~\ref{sec:error}.

\begin{table}[h]
\centering
\begin{tabular}{cccc}
\toprule
\# Tasks per workflow & \# Workflows & Pass Rate (\%) & Avg.\ time (min) \\
\midrule
1 & 37 & 48.6 & 13.1 \\
2 & 84 & 42.4 &17.4 \\
3 & 33 & 33.3 & 18.7 \\
$\geq$4 & 18 & 5.6  & 17.4 \\
\bottomrule
\end{tabular}
\caption{Average GPT 5.1 Pro completion time across workflows with different numbers of tasks.}
\label{tab:task_time}
\end{table}

We further analyze how GPT 5.1 web completion time scales with the number of tasks in a workflow (Table~\ref{tab:task_time}). The longest individual workflow run takes roughly 60 minutes to complete---but fails, highlighting how challenging workflows can be for current agents. On average, single-task workflows take 13.1 minutes, while workflows with two and three tasks require 17.4 and 18.7 minutes, respectively, reflecting the increased compositional complexity of multi-task settings. Interestingly, workflows with 4 or more tasks show a lower average time (17.4 minutes) than workflows with 3 tasks, most of which involve web search and almost all of which result in failure.

\begin{center}
  \includegraphics[width=\linewidth]{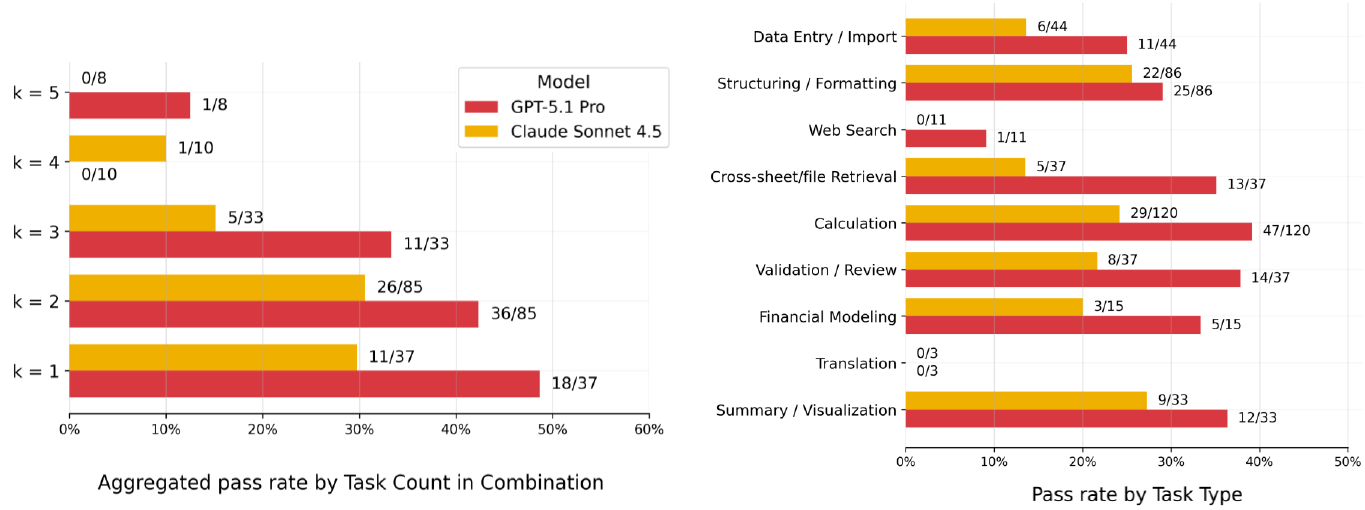}
  \captionof{figure}{Pass rate comparison for GPT 5.1 Pro and Claude Sonnet 4.5 across different task combinations and task types. The left chart visualizes the aggregated pass rate based on task combinations, revealing the models' capabilities in handling multi-step workflows commonly seen in professional finance and accounting tasks. The $k$ in the right chart represents the number of tasks included in a workflow. For example, a "k=3" workflow involves three distinct tasks, and its pass rate is calculated based on the collective performance of those tasks. The right chart shows the pass rates for individual tasks performed by both models in the \textsc{Finch} benchmark. For workflows that contain multiple tasks, a task is counted as correct only if the entire workflow is completed successfully. If a workflow fails, all tasks within that workflow are counted as incorrect. }
  \label{fig:task_breakdown}
\end{center}

In addition, we examine performance on a subset of 20 workflows that involve \textbf{non-spreadsheet artifacts} 
(e.g., PDFs, Word documents, and images). 
As shown in Table~\ref{tab:other_formats}, GPT~5.1~Pro achieves a 35.0\% pass rate on these workflows (7/20), lower than 
its overall rate of 38.4\%, while Claude Sonnet~4.5 achieves 25.0\% (5/20), 
matching its overall rate. This suggests that handling heterogeneous document 
types introduces additional challenges.

\begin{table}[h]
\centering
\small
\setlength{\tabcolsep}{4pt}
\begin{tabular}{lcc}
\toprule
\textbf{Pass Rate } & \textbf{Overall} & \textbf{Other file types} \\
\midrule
GPT-5.1 Pro (Product)        & 38.4\% & 35.0\% \\
Claude Sonnet 4.5 (Product)  & 25.0\% & 25.0\% \\
\bottomrule
\end{tabular}
\caption{Pass rates on workflows involving non-spreadsheet artifacts (e.g., PDFs, Word documents, and images) under human evaluation.}
\label{tab:other_formats}
\end{table}

Beyond binary pass/fail, our LLM-as-judge framework provides 
\textbf{rubric-level scores} along four dimensions: \emph{completeness}, 
\emph{correctness}, \emph{over-edit avoidance}, and 
\emph{readability}. Table~\ref{tab:rubric_scores} reports these 
scores for the two Claude product-side agents. Correctness is the primary bottleneck: Sonnet~4.5 achieves 
55.8\% completeness but only 36.0\% correctness.
These scores are produced by GPT-5-mini using structured 
output, where a single call assigns all four scores jointly; 
this may introduce inter-rubric confusion. More robust 
agentic evaluation methods are a worthwhile direction 
for future work.

\begin{table}[h]
\centering
\small
\setlength{\tabcolsep}{4pt}
\begin{tabular}{lcc}
\toprule
\textbf{Rubric (\%)} & \textbf{Sonnet 4.5} & \textbf{Opus 4.5} \\
\midrule
Completeness         & 55.8 & 69.8 \\
Correctness          & 36.0 & 45.3 \\
Over-edit Avoidance  & 55.8 & 64.0 \\
Readability          & 65.1 & 74.4 \\
\bottomrule
\end{tabular}
\caption{Rubric-level scores under automated evaluation.}
\label{tab:rubric_scores}
\end{table}

\paragraph{API-based.}
Figure~\ref{fig:intro_acc} further shows that the adapted API-based baselines are generally weaker than official product-side agents. Under automated evaluation, for example, GPT~5.1~Pro achieves a pass rate of 41.9\%, whereas GPT~5.1 with our API-based agent design reaches 32.0\%. A key limitation of the API-based baselines is that they rely on a single LLM call, which precludes iterative interaction, execution feedback, and self-correction—an important direction for future work in designing F\&A enterprise-grade agent frameworks.
Despite this constraint, our agent design narrows the performance gap by employing more efficient spreadsheet encodings and task-appropriate tool outputs within the single-call budget.

\subsubsection{Consistency Between Human and Automated Evaluation}

We adopt a lightweight multimodal model, GPT-5-mini, as the judge for automated evaluation framework. As shown in Table~\ref{tab:eval_results}, the automated evaluation largely aligns with human judgments: for GPT~5.1~Pro and Claude Sonnet~4.5, the judge agrees with human labels on 82.1\% and 90.2\% of workflows, respectively. The judge also achieves high recall (83.3\% and 88.4\%), meaning it recovers most human-labeled passes, and reasonably strong precision (73.6\% and 76.0\%), indicating that the majority of automatically predicted passes are also accepted by human evaluators. Overall, this suggests that automated evaluation may overestimate accuracy by several percentage points.

\begin{table}[h]
\centering
\caption{Comparison of human and automated evaluation on GPT and Claude product-side agents. ``Automated Eval'' shows pass counts/rates under the LLM-as-judge framework. ``Agreement w/ Human Eval'' reports how well the automated judgments match human labels: accuracy (Acc), recall (human-pass recall), and precision (human-pass precision).}
\label{tab:eval_results}
\scalebox{0.85}{
\begin{tabular}{lccccccc}
\toprule
\multirow{2}{*}{Model (Product)} &
\multicolumn{2}{c}{Automated Eval} &
\multicolumn{2}{c}{Human Eval} &
\multicolumn{3}{c}{Agreement w/ Human Eval} \\
\cmidrule(lr){2-3}\cmidrule(lr){4-5}\cmidrule(lr){6-8}
& Pass & Pass Rate (\%) & Pass & Pass Rate (\%) & Acc (\%) & Recall (\%) & Precision (\%) \\
\midrule
GPT 5.1 Pro       & 72/172 & 41.9 & 66/172 & 38.4 & 82.1 & 83.3 & 73.6 \\
Claude Sonnet 4.5 & 50/172 & 29.1 & 43/172 & 25.0 & 90.2 & 88.4 & 76.0 \\
\bottomrule
\end{tabular}
}
\end{table}

On the model side, the LLM judge can occasionally miss nuances in the rubric—either failing to catch subtle visual or numerical errors in large spreadsheets or, conversely, being overly literal about certain instructions (e.g., penalizing benign formula-to-constant conversions). However, we also observe cases where the LLM-based judge is correct but human raters are wrong---for example, when formulas are silently replaced with static values, which are difficult to detect through GUI-based inspection alone. On the system side, limitations of our spreadsheet tooling and data pipeline (e.g., incomplete support for corrupted but human-readable workbooks or uncommon file formats) can cause valid outputs to be marked as failures. Taken together, these factors mean that our automated scores should be interpreted as approximate rather than exact, and that human review remains important for borderline or high-impact workflows. 

\subsection{Error Analysis}
\label{sec:error}

To understand the sources of failure on \textsc{Finch}, we conducted a qualitative error analysis of GPT 5.1 Pro and Claude Sonnet 4.5 in both their product-agent and API configurations. For all failed workflows in our evaluation, we manually inspected the trajectories and annotated the primary cause of failure. 

From a workflow-centric perspective, we identify five dominant categories of error.\footnote{For this study, once we identify the first clear error in a failed workflow, we stop further analysis for that workflow.} Take Claude Sonnet 4.5 product-agent as an example. Across all examined failures, 10\% stem from \emph{task misunderstanding}: enterprise tasks often rely on implicit context in enterprise artifacts (e.g., spreadsheets), which models frequently overlook, leading them to misinterpret what is being asked and the required deliverable. 25\% are \emph{data retrieval errors}, including selecting the wrong cross-sheet, cross-table, or intra-table row/column ranges. 35\% arise from \emph{formula reasoning errors}, such as failing to reconstruct the latent business logic encoded in formulas or deriving incorrect new formulas. 25\% are due to \emph{code generation errors}, where generated scripts (e.g., Python with spreadsheet APIs) are syntactically invalid or misaligned with the spreadsheet layout. The remaining 5\% correspond to \emph{data rendering errors}, including incorrect formatting, misconfigured charts, or flawed final reports that deviate from the requested layout or narrative---for example, creating a brand-new spreadsheet instead of modifying the original one as requested. We also compare error patterns between web-based agents and API-based setups, with details provided in Appendix~\ref{error_analysis:details}.

Notably, all of these error types correspond to \emph{generic capabilities} that modern LLMs already appear to master on many existing benchmarks. The question, then, is why these ostensibly strong base abilities degrade so sharply on \textsc{Finch}. Our analysis points to five intertwined properties of real-world enterprise Finance \& Accounting workflows that make failures more likely and more catastrophic. First, \textsc{Finch} workflows routinely involve \emph{large, fragmented spreadsheet ecosystems}: dozens of interlinked workbooks and thousands of rows distributed across many sheets. Executing these workflows accurately requires long-range cross-sheet navigation and precise referencing, which substantially increases the likelihood of small retrieval errors. 
Second, the \emph{content is dense and semantically homogeneous}: many cells contain domain-specific financial concepts that are subtly different yet lexically similar (e.g., variants of revenue/expense items, adjusted vs. unadjusted metrics), making entity disambiguation and cell grounding unusually difficult. 
Third, the \emph{table layouts and structures are complex and often irregular}, including multi-level headers, merged cells, nested subtotals, and bespoke layouts that force the model to infer structure from noisy contents and ad hoc formatting. 
For example, at the code level, even tiny misinterpretations of these layouts (e.g., off-by-one errors when specifying ranges) can then propagate into globally incorrect outputs, especially when logic is applied in batch across many such sheets.
Fourth, \emph{formulas encode latent structure and logic}. In the \textsc{Finch} dataset, each sheet contains a large number of formulas that encode latent business logic, temporal assumptions, and fine-grained dependencies that are not visible from displayed values alone; yet models typically prioritize cell values and under-use formulas, leading to systematic misinterpretations. For example, in a pricing sheet with the column header \texttt{IF NGPL MidContinent index (@ Baker)}, the apparent semantics from the header alone suggest a daily exposure metric. However, inspecting the associated formula (\texttt{25 * V21 + C41 * C22}) reveals that the column in fact encodes a 55-day payment timing. Models that ignore or under-utilize formulas systematically misinterpret such columns’ roles in downstream calculations, and this misinterpretation then propagates through subsequent steps.
Finally, many workflows involve \emph{multimodal artifacts and chat-centric tasks} such as combining spreadsheets with PDFs, charts, and screenshots requiring the agent to jointly reason over heterogeneous formats. For example, tables embedded in PDFs are often only partially referenced, with key entries missing or truncated.


Many of these factors have been examined in prior work (e.g., multi-spreadsheet settings, complex table structures, formula reasoning, and multi-step workflows), and state-of-the-art models can perform reasonably well on benchmarks that emphasize a limited subset of these factors. In \textsc{Finch}, however, these factors co-occur within the same workflow in real-world enterprise data, and our results suggest this \textbf{composition} is what drives the sharp performance drop. \textsc{Finch} does not demand fundamentally new abilities; rather, it probes these abilities under an enterprise ``extreme'' regime of high complexity, noise, and long-horizon dependencies---closely mirroring real Finance \& Accounting work. Progress on compositional capability, therefore, requires training and evaluation on long-running, computation- and reasoning-intensive workflows over large, messy multimodal enterprise artifacts.

\section{Related Work}

The integration of LLMs into enterprise productivity tools has accelerated dramatically recently. The recently launched ChatGPT Agent~\citep{openai_agent} extends these capabilities to autonomous task completion, enabling multi-step workflows across web browsing, code execution, and spreadsheet manipulation. Microsoft Copilot~\citep{microsoft_copilot} embeds AI capabilities across the Microsoft 365 suite, enabling users to draft documents, analyze spreadsheets, and automate workflows through natural language interaction. Similarly, Google has integrated Gemini~\citep{gemini_workspace} into Google Workspace, providing AI-assisted features in Docs, Sheets, and Gmail. Anthropic's Claude Excel has also entered the enterprise space with spreadsheet automation capabilities~\citep{claude_excel}.

The emergence of agentic AI systems marks a significant shift from understanding and QA to autonomous task completion~\cite{li2023sheetcopilot,chen2024sheetagent,ma2024spreadsheetbench}. However, there are long-standing challenges such as messy inputs and multimodal processing. SpreadsheetLLM~\citep{spreadsheetllm2024} introduces novel encoding and compression methods to help LLMs understand large and messy spreadsheet structures, and further addresses this challenge through spreadsheet post-training. Beyond structural understanding, multimodal processing remains challenging for spreadsheet AI systems. On the formatting front, early work explored neural approaches for table formatting~\cite{dong2020neural}. Recent advances in formula generation have progressed from pretraining with numerical reasoning~\cite{cheng2022fortap} to natural language-driven formula synthesis~\cite{zhao2024nl2formula}, contrastive learning-based recommendation~\cite{chen2024auto}, and interactive formula prediction through hierarchical expansion~\cite{he2023hermes}. 

Recent years have seen significant progress in benchmarks for financial reasoning~\citep{chen2021finqa,islam2023financebench,finance_agent_benchmark,finben2024,findabench2025,finagentbench2025,finauditing2025,li2025investorbench,visfineval2025,gdpval2025,finsearchcomp2025,xfinbench2025,xie2024finben}, spreadsheet reasoning~\citep{li2023sheetcopilot,chen2024sheetagent,ma2024spreadsheetbench,sodbench2025,li2023auto,dong2019tablesense,sheetmind2025,spreadsheetllm2024,wu2025realhitbench,li2025mimotable,dong2025machinelearninglm,dong2025reasoning,li2024tapilot,zhao2024nl2formula,odysseybench2025,officebench2024}, and multimodal document~\citep{mathew2021docvqa}, table~\citep{li2020tablebank,zheng2024multimodal}, chart~\citep{wang2024charxiv}, table-chart~\citep{dong2024ttc}, and spreadsheet~\citep{xia2024vision,dong2019tablesense} reasoning, driving advances in LLM-based agents for enterprise tasks. However, \textsc{Finch} targets messy artifacts and long-horizon workflows in real-world finance and accounting settings.

\section{Conclusion}

We introduced \textsc{Finch}, a new benchmark for real-world F\&A enterprise workflows. \textsc{Finch} combines workflows induced from enterprise email threads, version histories of spreadsheets, and high-quality financial artifacts with rigorous expert annotation and a calibrated LLM-as-judge framework, enabling systematic evaluation of agents on diverse workflows that operate over large, messy, and multimodal enterprise artifacts and require long-horizon, spreadsheet-centric reasoning. Our experiments show that even the strongest frontier systems pass fewer than 50\% of workflows after spending 16.8 minutes per workflow, revealing a substantial gap between current AI capabilities and the demands of real enterprise practice. We hope \textsc{Finch} will serve as a foundation for developing agents to tackle real, messy and long-horizon professional work.

\section{Acknowledgments}
We thank Chi Zhang and Jieqiong Li for reviewing the dataset and providing valuable advice.

\clearpage
\bibliography{neurips_2025}
\bibliographystyle{plain}

\appendix

\newpage
\tableofcontents

\pagestyle{fancy}
\renewcommand{\headrulewidth}{0pt}
\fancyhead{}

\clearpage

\section{Author List}
\label{sec:author}

\begin{table}[h]
\centering
\small
\setlength{\tabcolsep}{6pt}
\renewcommand{\arraystretch}{1.35}
\begin{tabular}{@{}lll@{}}
\toprule
\textbf{Author} & \textbf{Affiliation} & \textbf{Email} \\
\midrule
Haoyu Dong\textsuperscript{*}  & UCAS; Microsoft              & \email{donghaoyu82@gmail.com} \\
Pengkun Zhang                  & South China University of Technology  & \email{sezhangpengkun@mail.scut.edu.cn} \\
Yan Gao                        & Zhongguancun Academy         & \email{gaoyan@zgci.ac.cn} \\
Xuanyu Dong                    & Harvest Fund                 & \email{qianmuxuanyu@126.com} \\
Yilin Cheng                    & Fudan University; Zhongguancun Academy     & \email{ylcheng23@m.fudan.edu.cn} \\
Mingzhe Lu                     & UCAS                         & \email{lumingzhe23@mails.ucas.edu.cn} \\
Zikun Zhu                      & Duke University              & \email{zz295@duke.edu} \\
Adina Yakefu                   & Hugging Face                 & \email{adina@huggingface.com} \\
Shuxin Zheng                   & Zhongguancun Academy         & \email{sz@bjzgca.edu.cn} \\
\bottomrule
\end{tabular}

\vspace{0.5em}
\noindent{\footnotesize \textsuperscript{*}Corresponding author. UCAS stands for the University of Chinese Academy of Sciences}
\end{table}

\section{Experiment Details}

\subsection{API-based Models}
\label{API-based:details}
\paragraph{Execution Paradigm.}
We frame the API evaluation as a \textbf{code generation task}. Models are instructed to solve spreadsheet manipulation and generation workflows by generating executable Python scripts, which are then executed in a sandboxed environment to produce output artifacts. This paradigm aligns with SpreadsheetBench's philosophy of treating model-written code as the primary action space, but is adapted here to accommodate long-horizon, multimodal enterprise tasks.

\begin{itemize}[leftmargin=*, itemsep=0.2em]
    \item \textbf{Action Space}: Models generate Python code using standard libraries including \texttt{openpyxl} (for Excel manipulation), \texttt{pandas} (for data processing), \texttt{matplotlib} (for visualization), and \texttt{scikit-learn} (for statistical analysis).
    \item \textbf{Output Format}: Models must produce complete, self-contained Python scripts wrapped in markdown code blocks (\texttt{```python ... ```}).
    \item \textbf{Sandboxed Execution}: Generated code is extracted via regex parsing and executed in isolated Docker containers running Jupyter Kernel Gateway. Each container mounts the dataset volume at \texttt{/mnt/data/} with a 10-minute session timeout.
    \item \textbf{Single-shot Protocol}: We employ a strict one-shot generation protocol without iterative refinement---each model produces exactly one solution per workflow. If the generated code fails to execute (e.g., due to syntax errors or runtime exceptions), the workflow is marked as failed without retry. This strict setting is designed to evaluate the model's raw code generation capability under realistic deployment constraints.
\end{itemize}

This unified code-as-action setting ensures that the measured performance reflects the model's inherent competence on complex workflows rather than benefits derived from interactive debugging.

\paragraph{Prompting Strategy.}
We employ a \textbf{zero-shot} setting with a structured system prompt comprising:
\begin{enumerate}[leftmargin=*, itemsep=0.2em]
    \item A role definition: ``You are an expert who can manipulate spreadsheets through Python code.''
    \item A detailed description of the compact spreadsheet encoding format with illustrative examples.
    \item The task instruction and explicit input/output file paths.
    \item Library-specific best practices (e.g., \texttt{openpyxl} chart creation patterns) to mitigate common code errors.
    \item An explicit directive to generate Python code as the final output.
\end{enumerate}

This structured design explicitly guides models toward generating valid, context-aligned Python code, minimizing ambiguity in task interpretation. However, for models that support reasoning traces (GPT 5.1, Gemini 3 Pro), we request explicit reasoning via the \texttt{include\_reasoning} API parameter, enabling us to capture the model's internal deliberation process for subsequent qualitative error analysis. Temperature is set to 0.7 across all models. 

\section{Detailed Analysis}
\label{error_analysis:details}
\paragraph{Web agents (ChatGPT 5.1 Pro vs. Claude Sonnet 4.5).} 
ChatGPT 5.1 Pro tends to decompose workflows into more, smaller steps, with explicit reasoning, tool calls, execution, and self-checking at each step. This leads to longer traces and noticeably higher latency, but also more opportunities for intermediate validation (e.g., sanity-checking partial results). However, the code it generates is often hidden behind tool abstractions, so our error attribution is limited to observed behavior and natural language reasoning rather than the exact implementation details.
Claude Sonnet 4.5 typically uses fewer steps and produces more direct solutions. In visualization-heavy workflows, its generated charts are often both more accurate and more aesthetically polished than those produced by ChatGPT 5.1 Pro, leading to relatively fewer failures in the data visualization sub-tasks.

ChatGPT 5.1 Pro and Claude Sonnet 4.5 agents can explore Excel files through many API calls within a single workflow, but their encoding methods are not well-suited to spreadsheets with complex layouts and structures. Thanks to efficient encoding and appropriate tool use, the following single-call API-based method achieves a pass rate that is much closer to that of product-side agents.

\paragraph{API-based}
Our API-based runs are single-call: they leverage the models’ underlying reasoning capabilities but lack two crucial affordances that web agents exploit. (i) interleaved code execution with feedback, and (ii) explicit reflection based on intermediate tool outputs. As a result, the API agents must generate the entire plan, code, and outputs within a single LLM call. When their initial structural assumptions about a spreadsheet are slightly off, they have no mechanism to detect or correct the mistake, leading to a significantly higher error rate, particularly in categories related to schema understanding and table manipulation. It's desirable for future work to explore agentic methods with multiple rounds of API calls.

\section{Tool-Call Overhead Case Study}
\label{appendix:tool_call}

To better understand the practical complexity of completing 
\textsc{Finch} tasks, we conduct a small-scale case study on 
tool-call overhead. From the full set of 172 workflows, we select 
20 representative examples covering different task types and varying 
levels of complexity, and analyze how many tool calls are required 
for an LLM-based agent to complete them.

\subsection{Methodology}

For each selected task, we use the up-to-date (March 2026) Claude Coworker (Opus~4.6) in an 
isolated session and ask it to solve the task according to the 
provided instruction. The agent is also required to write its 
tool-use trajectory into a log file, recording each tool call 
step by step. Since the number of tool calls in a failed attempt 
may differ from that in a successful completion, we use a second 
session to compare the produced output against the reference 
answer file and determine whether the result is correct.

When the output is incorrect, we observe two distinct cases from 
the perspective of tool usage. In the first case, the overall 
tool-use procedure is already complete, but the execution within 
one or more steps is wrong; for example, the agent invokes a 
Python script as needed, but the script contains an incorrect 
formula. In the second case, the trajectory is missing one or 
more critical steps that are necessary for producing the correct 
output. This second case is relatively uncommon, appearing only 
twice among the 20 examples we analyze. For such cases, we ask 
the model to identify the missing indispensable steps and merge 
them into the original trajectory. Therefore, our final count is 
intended to approximate the tool calls required to correctly 
complete the task, rather than merely reporting the raw tool calls 
from a failed run.

\subsection{Results and Analysis}

Table~\ref{tab:tool_call_case_study} summarizes the 20 selected 
tasks and their corresponding numbers of tool calls.

\begin{table}[h]
\centering
\small
\begin{tabular}{l l l l}
\toprule
Workflow ID & Task Type(s) & \#Tasks & \#Tool Calls \\
\midrule
46  & Calculation & 1 & 7 \\
31  & Summary / Visualization & 1 & 8 \\
26  & Summary / Visualization & 1 & 10 \\
47  & Calculation & 1 & 13 \\
17  & Cross-sheet/file Retrieval & 1 & 14 \\
137 & Structuring / Formatting & 1 & 16 \\
91  & Calculation, Financial Modeling & 2 & 6 \\
106 & Structuring / Formatting, Calculation & 2 & 6 \\
49  & Validation / Review, Calculation & 2 & 14 \\
16  & Calculation, Summary / Visualization & 2 & 17 \\
28  & Validation / Review, Structuring / Formatting & 2 & 17 \\
158 & Validation / Review, Calculation & 2 & 18 \\
15  & Structuring, Translation, Summary / Vis. & 3 & 9 \\
150 & Calculation, Data Entry, Cross-sheet Retrieval & 3 & 9 \\
66  & Calculation, Structuring, Financial Modeling & 3 & 17 \\
139 & Financial Modeling, Calculation, Summary / Vis. & 3 & 25 \\
159 & Structuring, Calc., Fin.\ Modeling, Validation & 4 & 14 \\
81  & Data Entry, Structuring, Validation, Calc. & 4 & 17 \\
19  & Web Search, Data Entry, Structuring, Calc. & 4 & 107 \\
88  & Web Search, Data Entry, Structuring, Calc., Vis. & 5 & 71 \\
\bottomrule
\end{tabular}
\caption{Case study of tool-call overhead on 20 representative tasks.}
\label{tab:tool_call_case_study}
\end{table}

\paragraph{Web-grounded tasks incur substantially higher tool-call overhead.}
Among the 20 cases, the two tasks involving web search (ID~19 
and ID~88) require dramatically more tool calls than the others. 
In Task~19, \texttt{websearch} and \texttt{webfetch} together 
account for 60 out of 107 tool calls (56\%). In Task~88, these 
two tools account for 34 out of 71 tool calls (48\%). During 
inspection, we find that web-grounded tasks often require repeated 
retrieval from multiple external sources, followed by cross-source 
comparison, integration, and verification. As a result, the 
dominant cost in such tasks comes not from spreadsheet manipulation 
itself, but from external information acquisition and validation.

\paragraph{Distribution of tool calls for non-web tasks.}
After excluding the two web-search tasks, the number of tool calls 
ranges from 6 to 25, with a mean of 13.2 and a median of 14. 
The first and third quartiles are 9 and 17, respectively, giving 
an interquartile range of 8.

\paragraph{Task-type count alone does not explain tool-call overhead.}
After removing the two web-search outliers, we do not observe a 
clear positive relationship between the number of task-type labels 
and the number of tool calls. For example, both Task~46 and 
Task~47 are labeled as \textit{Calculation}, but Task~47 requires 
nearly twice as many tool calls due to additional business-logic 
reasoning (sign conventions, Purchase vs.\ Pooling Method 
tax-base logic, and cross-checking of business consistency).

\paragraph{Limitations.}
This analysis should be viewed as a case study rather than a 
full-scale statistical evaluation. The 20 tasks are selected as 
representative examples rather than a random sample, and the 
measured tool-call counts depend on the behavior of a specific 
agent configuration (Claude Coworker, Opus~4.6). In addition, 
our corrected trajectories for failed runs involve a post hoc 
reconstruction of missing indispensable steps, which introduces 
some degree of judgment into the counting process.

\subsection{Representative Tool-Call Logs}
\label{appendix:tool_call_logs}

We present two representative tool-call logs.
Task~26 illustrates a standard successful trajectory;
Task~139 illustrates a case where the original trajectory
omits critical steps (marked with
\addedtool{Bash}) that must be added for correct completion.

\subsubsection{Task 26: Chart Generation across Scenario Sheets}
\label{appendix:task26_tool_calls}

\paragraph{Task.}
Add two charts to each of the three worksheets
(\texttt{Scenario1--3}):
(1)~a \textbf{Hedge Coverage} stacked column chart
    (first year; X = months; stacked series = contract MW;
     Short = reference line; Y = MW), and
(2)~a \textbf{Portfolio Cost} line chart
    (all years; X = months; series = Portfolio Weighted Avg
     and Monthly Price; Y = \$/MWh).

\begin{tcolorbox}[
  colback=logbg, colframe=logframe,
  boxrule=0.4pt, arc=2pt,
  left=4pt, right=4pt, top=4pt, bottom=4pt,
  fonttitle=\small\sffamily\bfseries,
  title=Tool-call log\; (10 calls)]
\begin{enumerate}[leftmargin=1.6em, itemsep=0.15em,
                   label={\small\sffamily\bfseries\color{gray}\arabic*.}]
\item \tool{Bash} \texttt{ls mnt/26/}\,---\,confirm source file exists.
\item \tool{Read} \texttt{metadata.json}\,---\,parse task instructions and output requirements.
\item \tool{Read} \texttt{SKILL.md}\,---\,review best practices for Excel operations.
\item \tool{Bash} Python: inspect workbook structure via \texttt{openpyxl} (sheet names, first 20 rows, column mapping).
\item \tool{Bash} Read full output (previous step was truncated).
\item \tool{Bash} Python: inspect first 20 rows of each Scenario sheet in detail---column semantics, data types.
\item \tool{Bash} Python: read date column (Z, rows 4--63) to identify first-year range (rows 4--15) and full range (Jan~2001--Dec~2005).
\item \tool{Bash} \texttt{cp 26\_src\_0.xlsx 26.xlsx}\,---\,preserve original, work on copy.
\item \tool{Bash} Run \texttt{add\_charts.py}: create all 6 charts (2 per Scenario sheet) via \texttt{openpyxl}.
\item \tool{Bash} Python: verify chart count, titles, types, and series per sheet.
\end{enumerate}
\end{tcolorbox}

\subsubsection{Task 139: XNPV Sensitivity Analysis with Corrected Trajectory}
\label{appendix:task139_tool_calls}

\paragraph{Note.}
This is a \emph{corrected} tool-call log.
Steps marked \addedtool{Bash} were missing from the
original trajectory and added post hoc; they mainly involve
constructing monthly cash-flow data for \texttt{XNPV Variables},
building multi-year analysis data for \texttt{Cedar Brakes},
and writing native Excel \texttt{XNPV} formulas.

\begin{tcolorbox}[
  colback=logbg, colframe=logframe,
  boxrule=0.4pt, arc=2pt,
  left=4pt, right=4pt, top=4pt, bottom=4pt,
  fonttitle=\small\sffamily\bfseries,
  title=Tool-call log\; (25 calls),
  breakable]

\phase{Phase 1 --- Understand the workbook}
\begin{enumerate}[leftmargin=1.6em, itemsep=0.15em,
                   label={\small\sffamily\bfseries\color{gray}\arabic*.}]
\item \tool{Bash} \texttt{ls mnt/139/}\,---\,confirm source file.
\item \tool{Read} \texttt{metadata.json}\,---\,task instructions, source file list, output requirements.
\item \tool{Read} \texttt{SKILL.md}\,---\,Excel operation best practices.
\item \tool{Bash} Python (\texttt{data\_only=True}): inspect all sheet names, dimensions, and data values.
\item \tool{Bash} Python: read \texttt{Damage Calculations} sheet---annual capacity payments (Q1--Q4), XNPV values, sensitivity-table region.
\item \tool{Bash} Python: read \texttt{Summary} sheet---layout of Q1--Q4 sections and Delta region.
\item \tool{Bash} Python (\texttt{data\_only=False}): read formulas in \texttt{Damage Calculations}---calculation logic, \texttt{DataTableFormula} objects.
\item \tool{Read} Read full output of step~7 (truncated).
\item \tool{Bash} Python: read formulas in \texttt{Summary} sheet---formula reference relationships.
\item \tool{Bash} Python: read formulas and variables in \texttt{Tenaska's i~Adjustment} sheet.
\item \tool{Bash} Python: read variable values in \texttt{Tenaska's i~Adjustment}.
\item \tool{Bash} Python: read formulas and variables in \texttt{Brazos' i~Adjustment}---compare with Tenaska.
\item \tool{Bash} Python: validate XNPV calculation method---date conventions, discounting logic.
\item \tool{Bash} Python: inspect \texttt{DataTableFormula} parameters---confirm input variable is \texttt{F8} (plant capacity).
\end{enumerate}

\phase{Phase 2 --- Build missing intermediate data \normalfont\itshape(added steps)}
\begin{enumerate}[leftmargin=1.6em, itemsep=0.15em,
                   label={\small\sffamily\bfseries\color{gray}\arabic*.},
                   resume]
\item \addedtool{Bash} Python: read \texttt{Cedar Brakes} sheet---annual data layout (\texttt{Year}, \texttt{Volume}, \texttt{Old PPA}, \texttt{Market}, \texttt{SI}, \texttt{New PPA}, \texttt{Savings}), column relationships, year range (2000--2016+), and rows/columns to fill.
\item \addedtool{Bash} Python: compute annual capacity payments for Q1--Q4 under four capacity assumptions (240/245/258/263 MW), applying the rate structure and interest-rate adjustments from Tenaska and Brazos.
\item \addedtool{Bash} Python: convert annual payments into monthly cash-flow sequences (date sequence from Jan~1997 to contract end; monthly = annual/12); verify totals.
\end{enumerate}

\phase{Phase 3 --- Execute main script}
\begin{enumerate}[leftmargin=1.6em, itemsep=0.15em,
                   label={\small\sffamily\bfseries\color{gray}\arabic*.},
                   resume]
\item \tool{TodoWrite} Create task list.

\item \tool{Bash} Main Python script---copy source to \texttt{139.xlsx}, then:

\smallskip
\hspace{0.5em}%
\begin{minipage}[t]{0.92\linewidth}
\addedtool{XNPV Variables}\;
Write title, headers (\texttt{Date}, \texttt{Q1--Q4}), and monthly cash-flow rows starting at row~88 of \texttt{Damage Calculations}; scale payments proportionally for 240/245/258/263~MW.

\smallskip
\addedtool{Cedar Brakes}\;
Fill multi-year financial data (rows~24+): \texttt{Volume}, \texttt{Old/New PPA}, \texttt{Market}, \texttt{SI}, \texttt{Savings} for all years 2000--end.

\smallskip
\addedtool{XNPV formulas}\;
Write \texttt{=XNPV(rate, flows, dates)} formulas into the sensitivity-table region, referencing the cash-flow data above.

\smallskip
\tool{Original}\;
Fill Q1--Q4 capacity values and Delta section (differences vs.\ 263~MW baseline) in \texttt{Summary}; save.
\end{minipage}

\item \tool{TodoWrite} Update progress.
\item \tool{Glob} Locate \texttt{recalc.py}.
\item \tool{Bash} \texttt{python recalc.py}\,---\,recalculate all Excel formulas (including XNPV).
\end{enumerate}

\phase{Phase 4 --- Verify outputs}
\begin{enumerate}[leftmargin=1.6em, itemsep=0.15em,
                   label={\small\sffamily\bfseries\color{gray}\arabic*.},
                   resume]
\item \addedtool{Bash} Python (\texttt{data\_only}): verify \texttt{XNPV Variables} integrity (row count, date continuity, cash-flow values), \texttt{Cedar Brakes} data, and XNPV formula results.
\item \tool{Bash} Python: verify \texttt{Summary} (PV5, debt, SWAP, implied equity for Q1--Q4) and \texttt{Damage Calculations} sensitivity table (all XNPV values, Delta vs.\ 263~MW baseline).
\item \tool{TodoWrite} Final progress update.
\end{enumerate}

\end{tcolorbox}

\clearpage

\section{Ethics Statement}

The \textsc{Finch} benchmark is constructed entirely from existing, publicly available data sources. Concretely, our workflows are derived from (1) the Enron email corpora, including the parsed Enron email dataset on Kaggle (released under the CC0 Public Domain dedication) and the Enron Email Dataset from EnronData.org (licensed under CC BY 3.0 US); (2) the EUSES spreadsheet corpus and its modified variants (CC BY 4.0); and (3) a diverse collection of enterprise-like artifacts, including documents from investment and securities companies, the World Bank (CC BY 3.0), Canadian and British government websites (Open Government License), and public corpora such as WideSearch (MIT license) and DABStep (CC BY 4.0). We respect the original licenses of all upstream resources and only redistribute content within the terms they allow.

On top of these sources, we apply additional filtering, normalization, and expert annotation to organize spreadsheets and related documents into coherent workflows with task instructions, input files, and reference outputs. We do not introduce any new personally identifiable information. During curation, we remove obviously sensitive fields when they are not necessary for the task (e.g., personal contact information or signatures) and avoid annotating workflows whose successful completion would depend on sensitive personal attributes rather than business logic. The resulting \textsc{Finch} dataset is released under the Creative Commons Attribution 3.0 United States license (CC BY 3.0 US), which permits broad reuse while requiring appropriate attribution.

The language in \textsc{Finch} is primarily English, reflecting the dominant language of the underlying Enron and EUSES corpora and many of the public institutional sources. Because some artifacts originate from funds and securities institutions and from Canadian government materials, a small fraction of workflows include Chinese or French content.


\section{Examples}
\label{sec:example}

This section presents representative examples from \textsc{Finch} to illustrate the diversity and complexity of spreadsheet-centric enterprise workflows covered by the benchmark. The examples span a wide range of realistic professional tasks, including cross-sheet verification, formula auditing, document-grounded extraction, schema transformation, financial modeling, and reporting.

Note that the figures shown in this section are illustrative excerpts rather than complete workflow inputs or full task specifications. Many \textsc{Finch} workflows involve large, multi-sheet or multi-file inputs and extensive supporting documents, which cannot be fully visualized in static screenshots. Accordingly, the figures are intended to highlight representative data characteristics, structural complexity, and reasoning challenges that arise in the underlying workflows, rather than to fully specify the tasks themselves.

Across these workflows, AI agents are required to jointly reason over heterogeneous artifacts, while preserving structural consistency and semantic correctness. Many tasks demand multi-step reasoning, precise interpretation of formulas, and careful cross-referencing across sheets or external documents. Collectively, these examples highlight the core challenges targeted by \textsc{Finch}: messy and multimodal document processing, long-horizon reasoning, robust code generation under schema variation, and faithful execution of complex financial and accounting logic.

\subsection{Example 1}
\begin{center}
  \includegraphics[width=\linewidth]{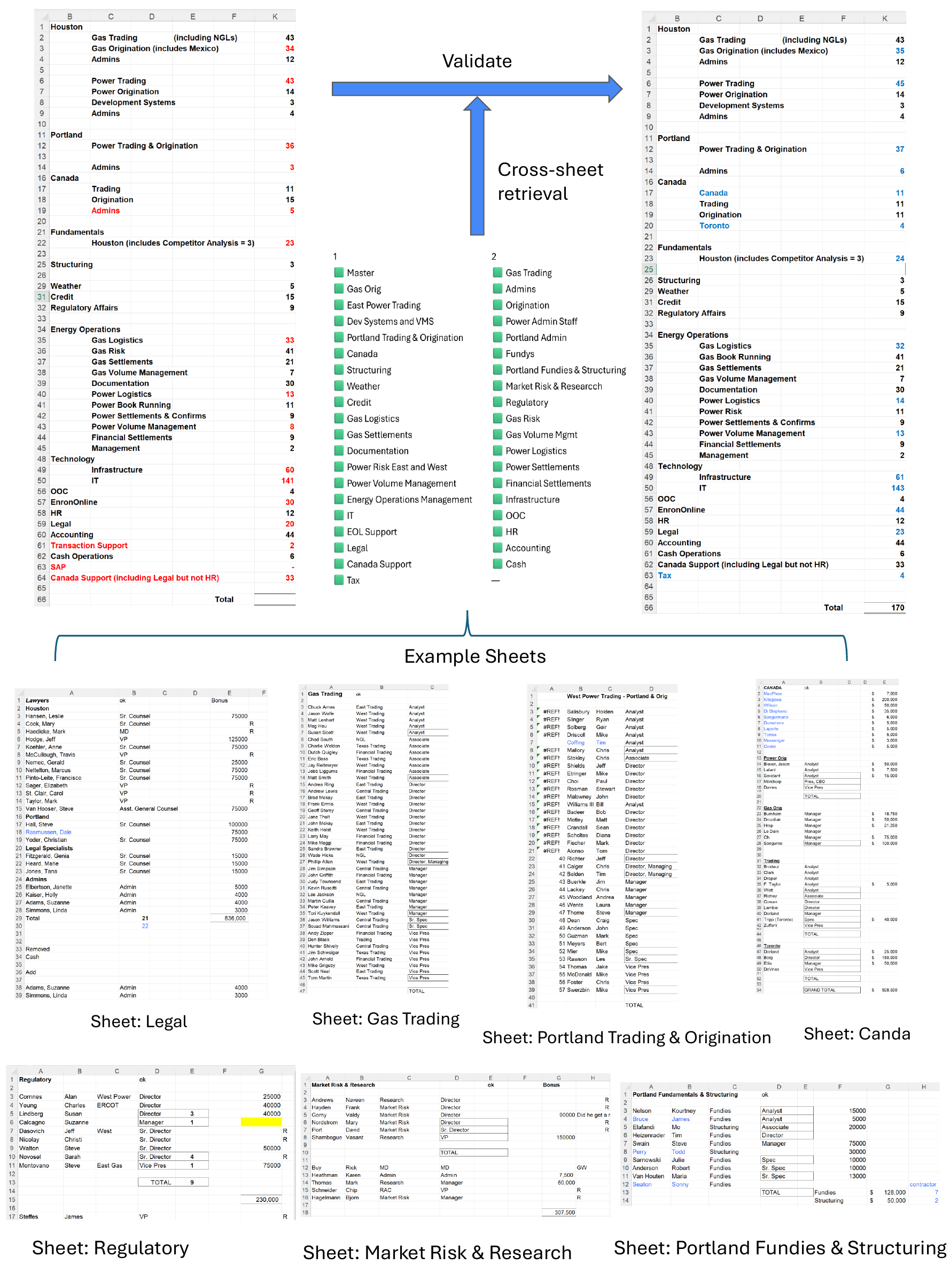}
  \captionof{figure}{For this task, the model must verify the department headcount summary by cross-checking each of the \textbf{39} departments against its detailed roster sheet. It should correct discrepancies such as miscounts and missing or duplicate entries. The summary must be updated by fixing incorrect totals, removing departments that no longer exist, and adding any omitted departments. Furthermore, the underlying schema varies slightly across departments, which challenges reliable code generation.}\label{fig:example_yan_1}
\end{center}

\subsection{Example 2}

\begin{center}
  \includegraphics[width=\linewidth]{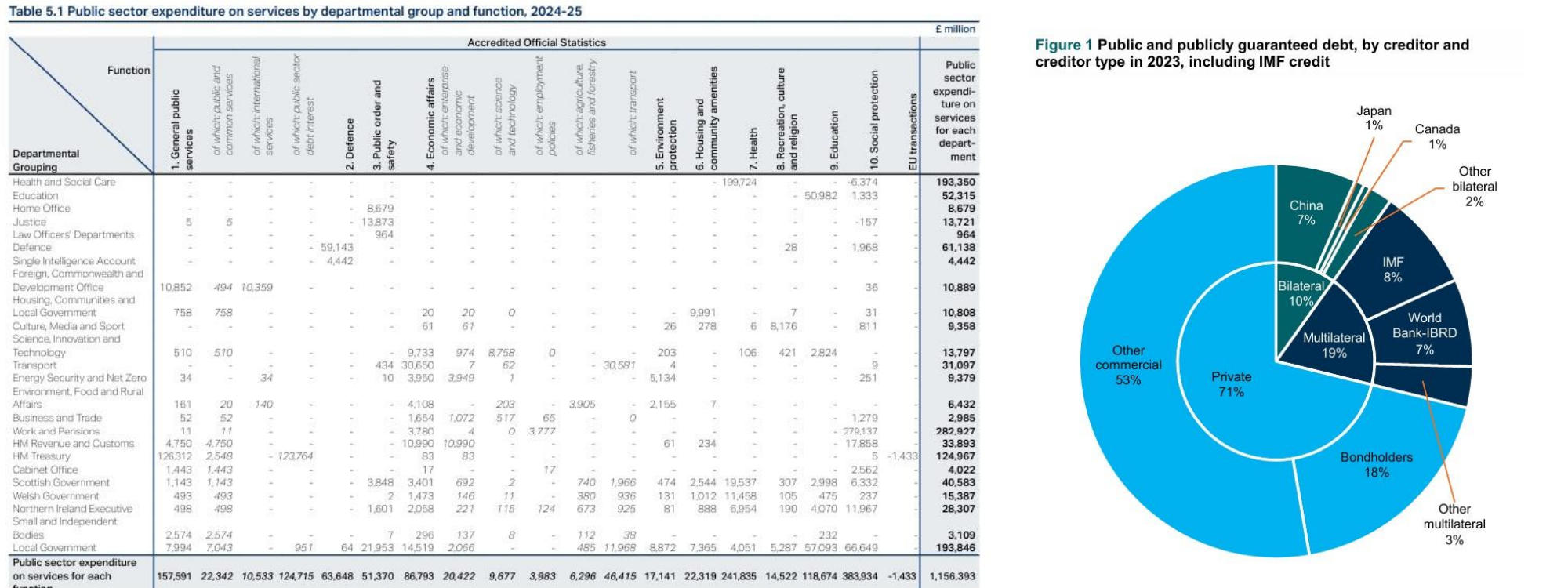}
  \captionof{figure}{An example of extracting data from tables and charts in PDFs and saving it to a spreadsheet. AI agents must understand the layout, parse hierarchical structures, interpret how values map to specific cells, and reconstruct formulas from aggregated values.
 }\label{fig:hy_example_9}
\end{center}

\subsection{Example 3}

\begin{center}
\includegraphics[width=\linewidth]{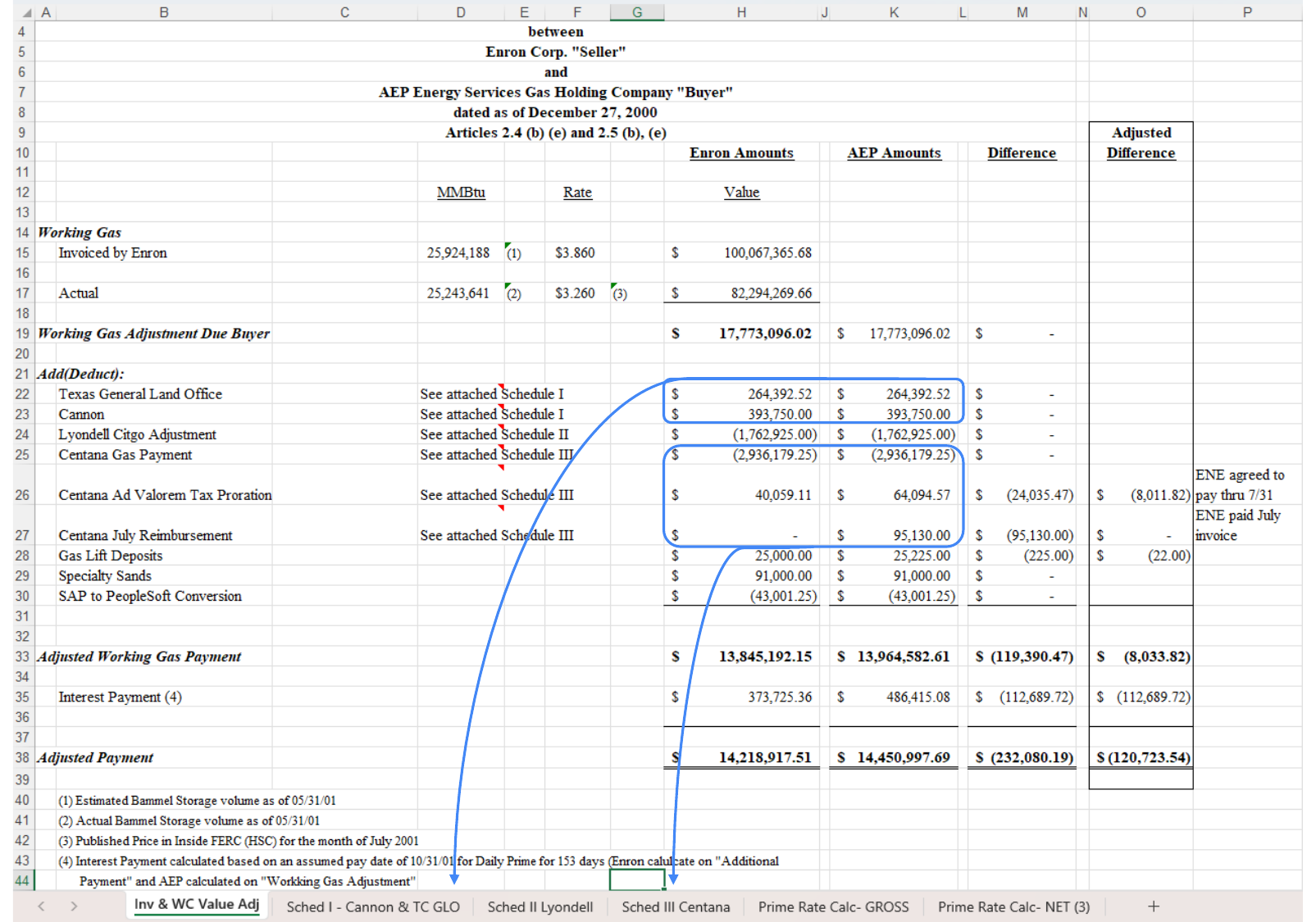}
  \captionof{figure}{Cross-sheet reference validation. This example is relatively easy for frontier AI agents.
}\label{fig:hy_example_6}
\end{center}

\clearpage

\subsection{Example 4}
\begin{center}
  \includegraphics[width=\linewidth]{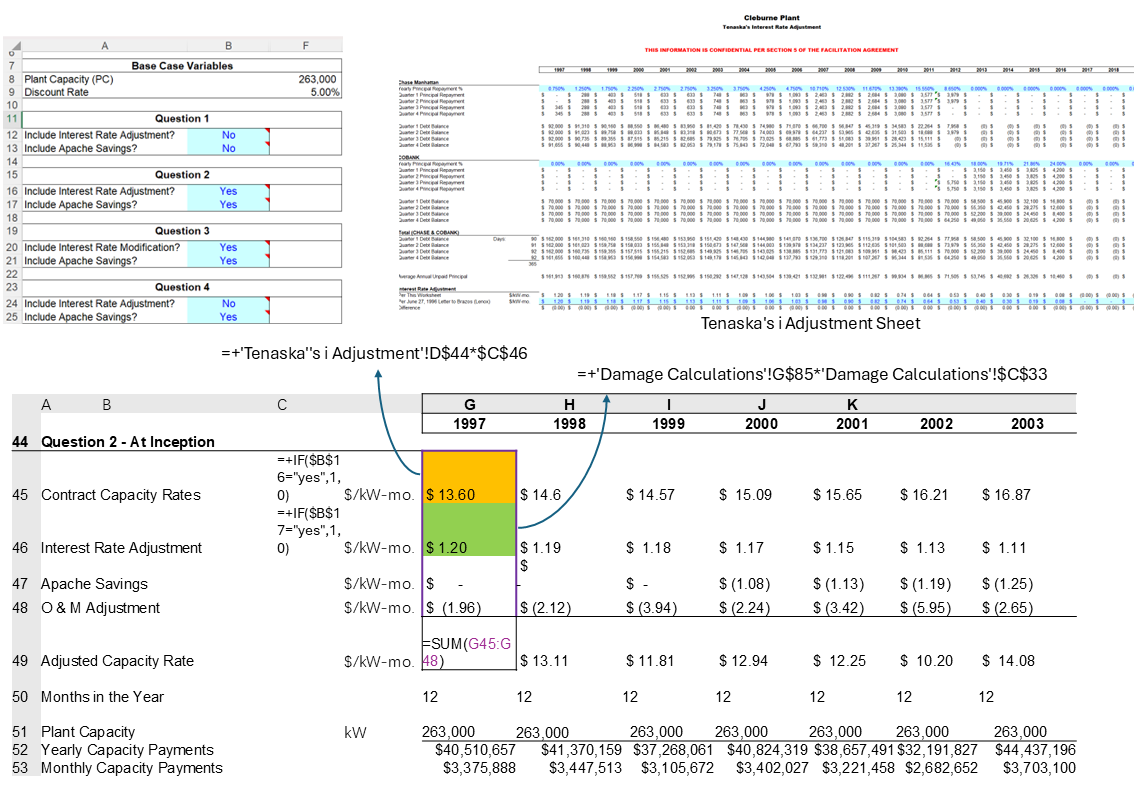}
  \captionof{figure}{This task requires deriving the XNPV5 of the contract under different combinations of assumptions. The analysis uses contract capacity rates, plant capacity, and the specified discount rate provided in the table. While key adjustment components---namely the Interest Rate Adjustment, Apache Savings, and O\&M Adjustment---must be retrieved from supporting documents and applied according to each scenario (included, excluded). 
  For each assumption set, the analyst must then construct the annual capacity payment cash flows by deriving the adjusted capacity rate, converting it into monthly and annual capacity payments, and assembling the full month-by-month cash-flow schedule. Only after these intermediate steps are completed can the cash flows be discounted to the valuation date (e.g., December 31, 2000) to compute XNPV5.} \label{fig:example_yan_3}
\end{center}

\subsection{Example 5}

\begin{center}
  \includegraphics[width=\linewidth]{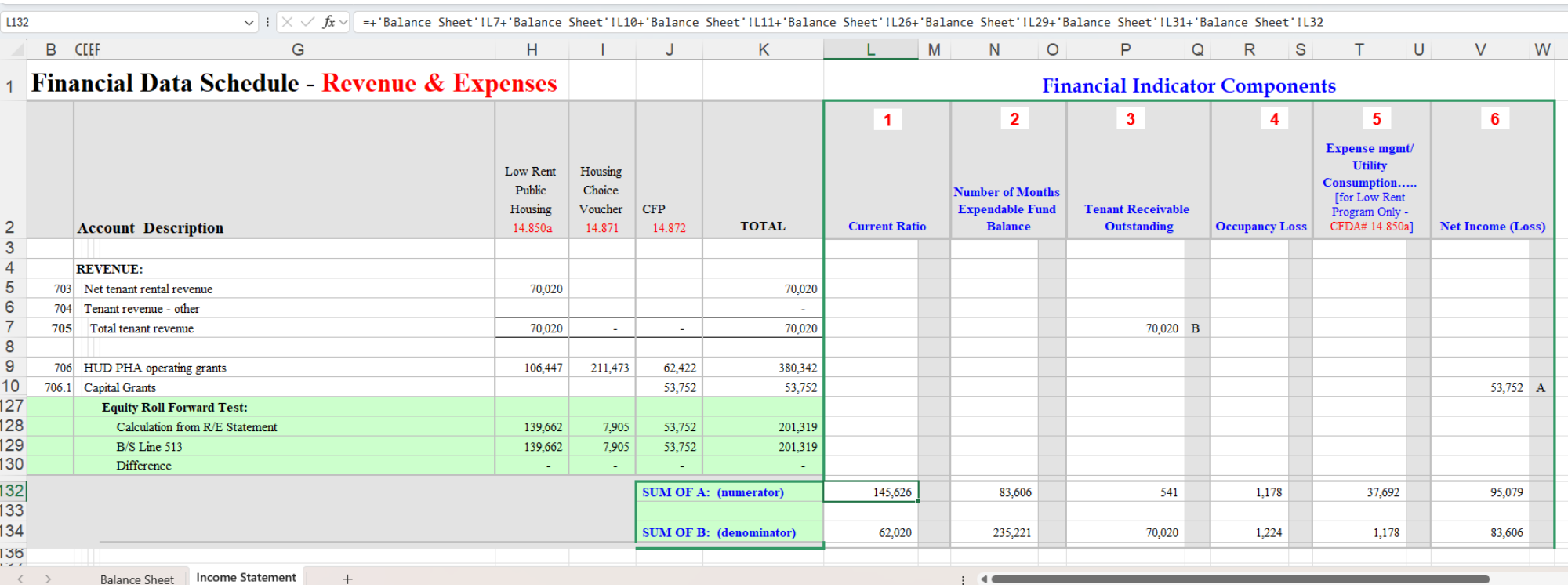}
  \captionof{figure}{The sum of A\&B and the equity roll-forward test require cross-sheet retrieval and calculation.
}\label{fig:hy_example_8}
\end{center}

\subsection{Example 6}

\begin{center}
  \includegraphics[width=\linewidth]{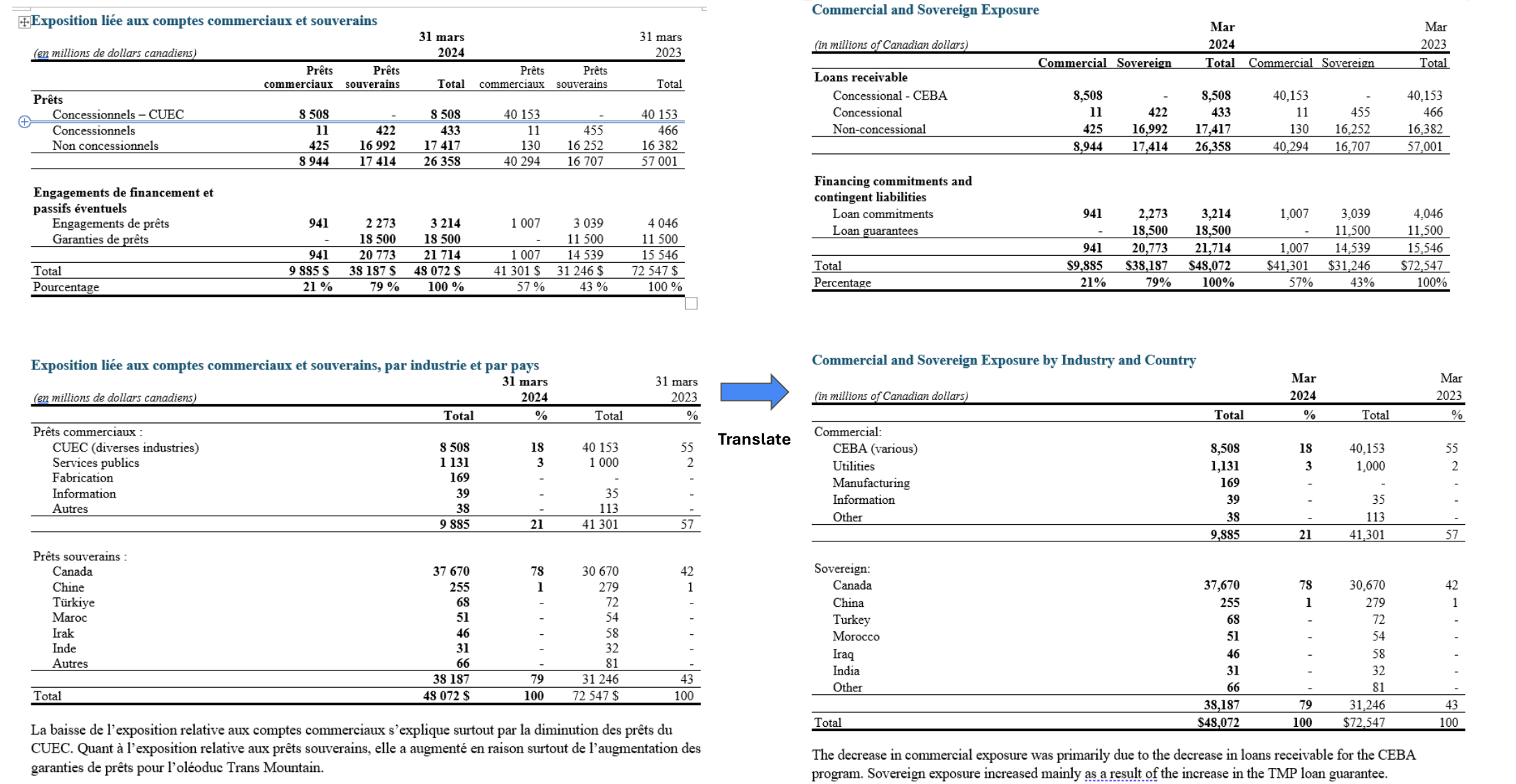}
  \captionof{figure}{A workflow that translates a French report into English while preserving its format and structure. The report contains many tables to translate, along with text, notes, and even charts.
 }\label{fig:hy_example_11}
\end{center}

\subsection{Example 7}

\begin{center}
  \includegraphics[width=\linewidth]{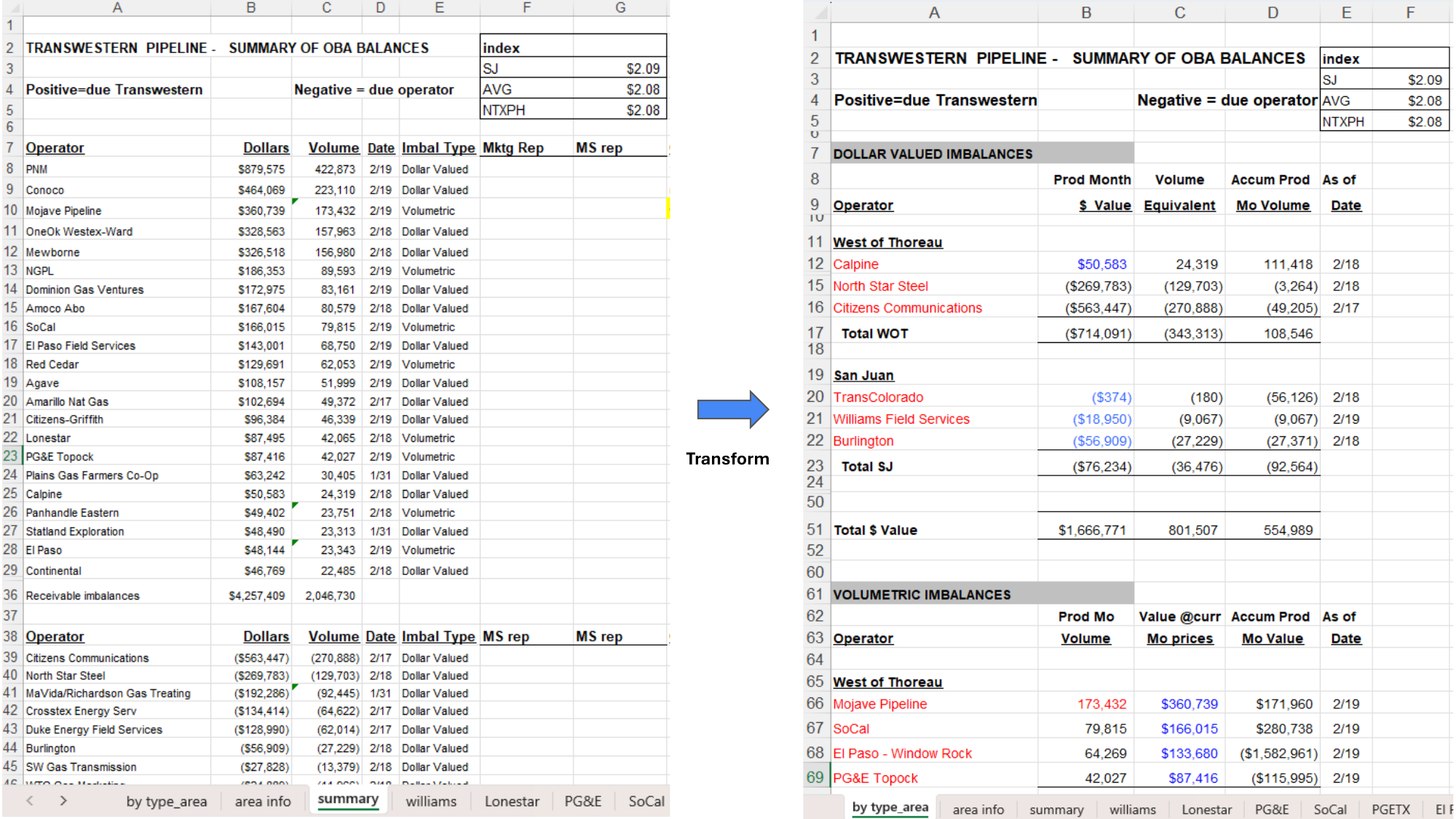}
  \captionof{figure}{Transforming a table from one structure to another requires reorganizing data and retrieving information across sheets (e.g., \texttt{area\_info} and \texttt{summary}). This example poses additional challenges: (1) distinguishing value-driven operators from volume-driven operators, and (2) performing aggregation and validation over the reorganized data.
}\label{fig:hy_example_7}
\end{center}

\subsection{Example 8}
\begin{center}
  \includegraphics[width=\linewidth]{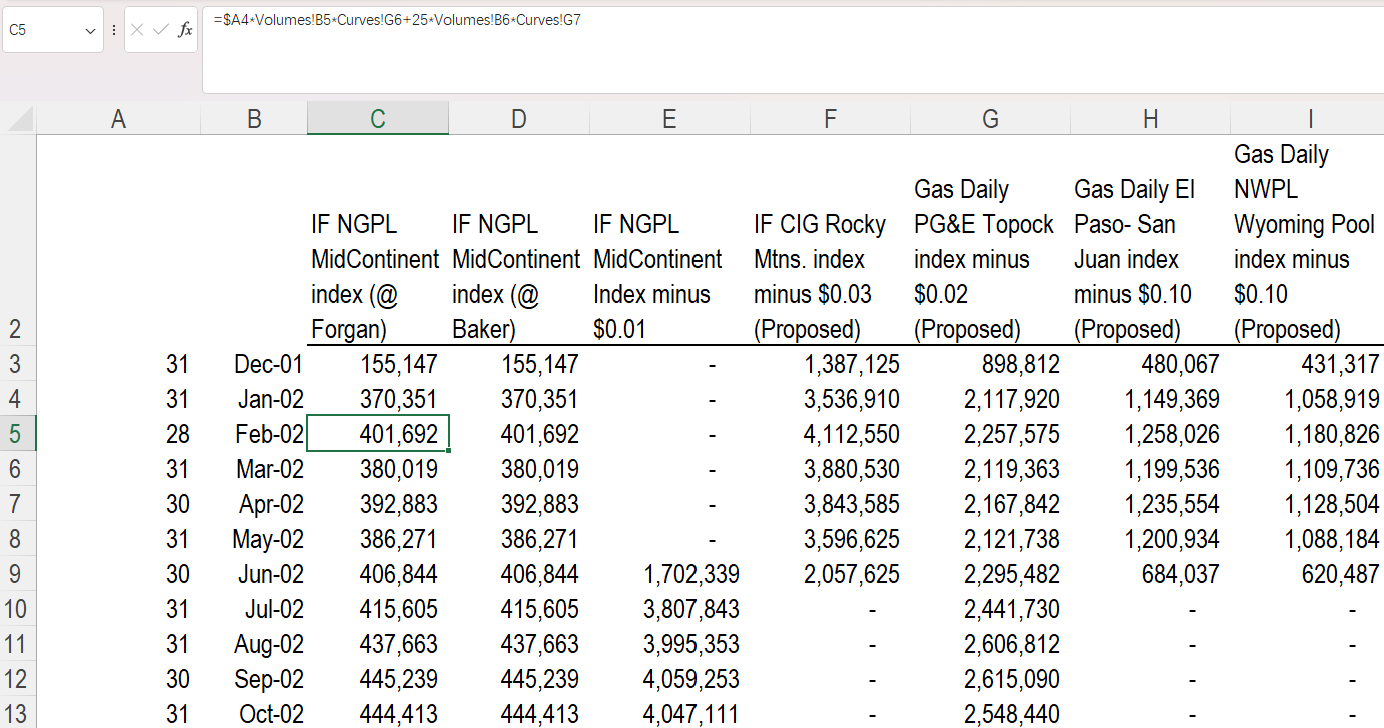}
  \captionof{figure}{The apparent semantics from the headers suggest a monthly/daily exposure metric. However, inspecting the underlying formula (e.g., \texttt{C5=\$A4*Volumes!B6*Curves!G7+25*Volumes!B7*Curves!G8}) reveals that it actually encodes a 55-day payment timing schedule. Models that ignore or underutilize formula information, therefore systematically misattribute the column’s role in downstream computations, and this misinterpretation then propagates through subsequent steps.}\label{fig:example_yan_2}
\end{center}

\subsection{Example 9}

\begin{center}
\includegraphics[width=\linewidth]{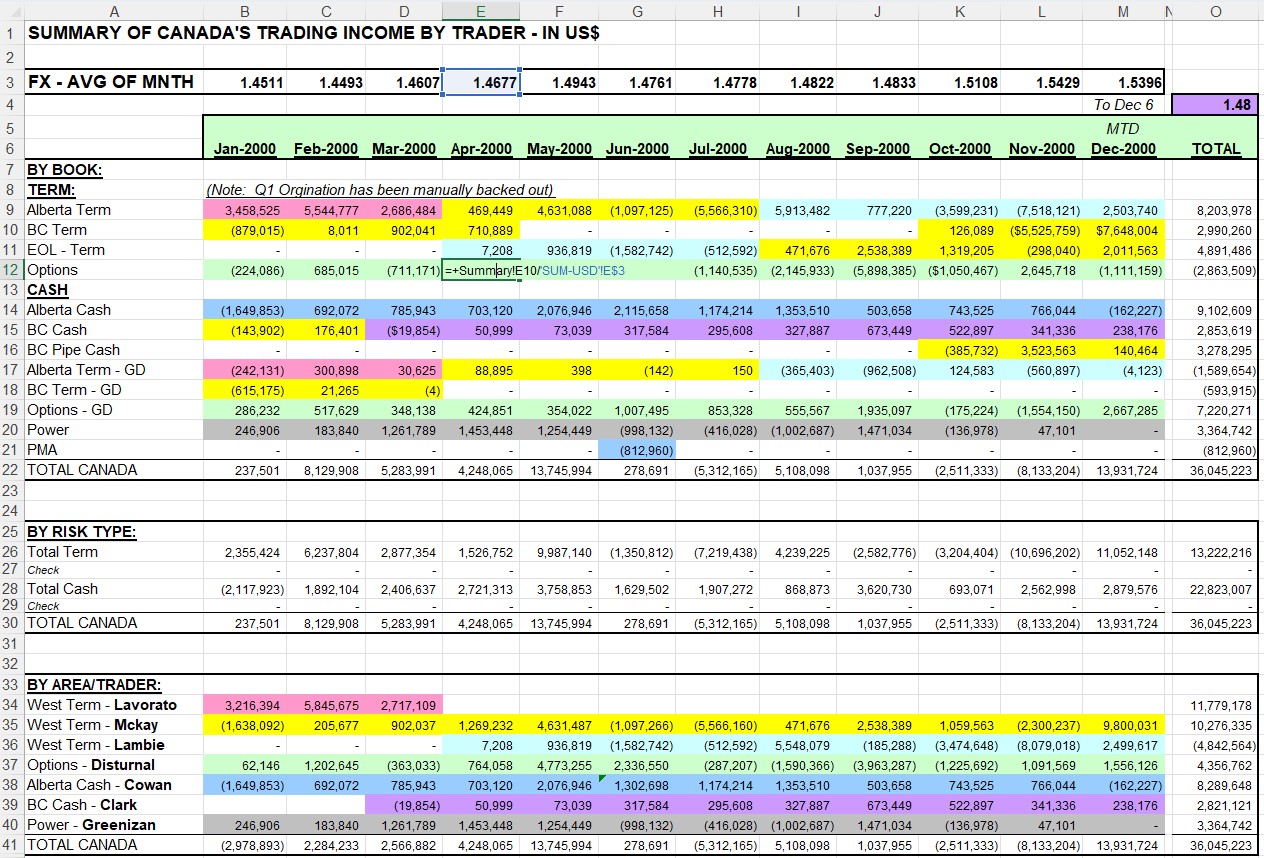}
  \captionof{figure}{This workflow requires creating a new spreadsheet with all values converted to USD. It also requires correct in-sheet and cross-sheet formula references while preserving the original spreadsheet layout.}\label{fig:hy_example_2}
\end{center}

\subsection{Example 10}

\begin{center}
\includegraphics[width=\linewidth]{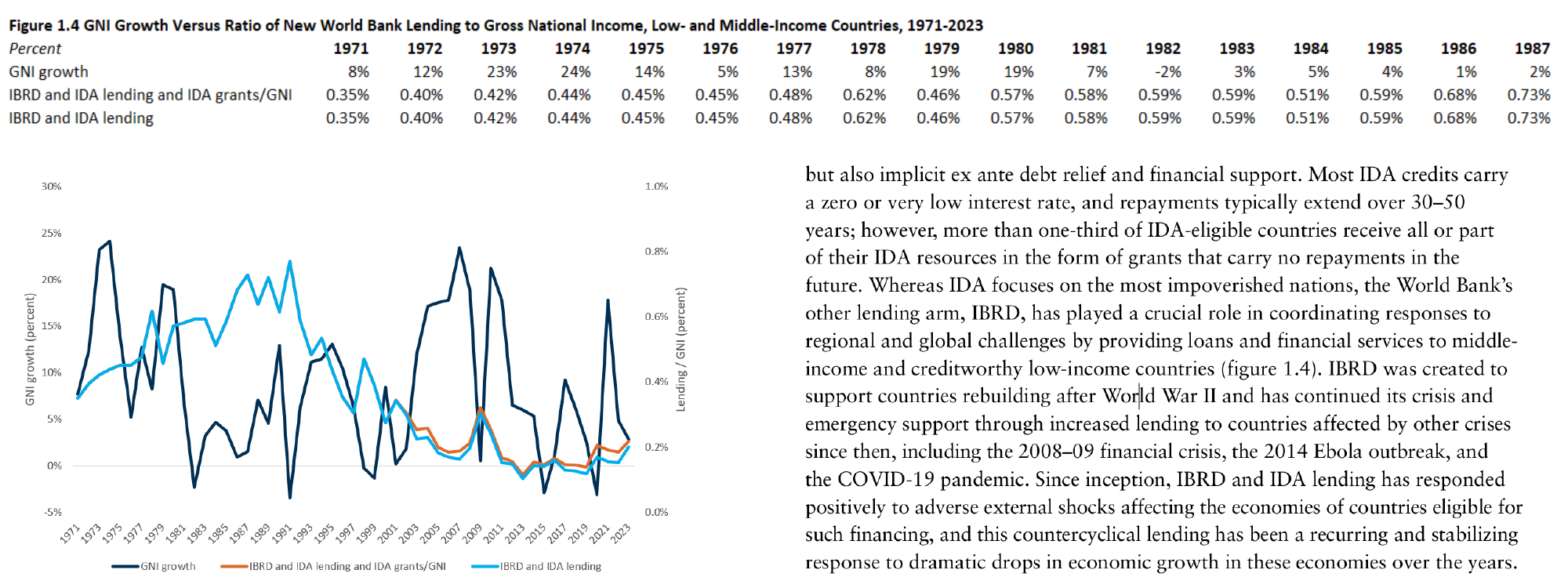}
  \captionof{figure}{Generating reports from tabular data requires financial knowledge of data analysis, financial events, and visualization. For example, one may plot two series with different units on a single chart (e.g., using a secondary y-axis) to reveal their correlation.}\label{fig:hy_example_10}
\end{center}

\subsection{Example 11}
\begin{center}
\includegraphics[width=\linewidth]{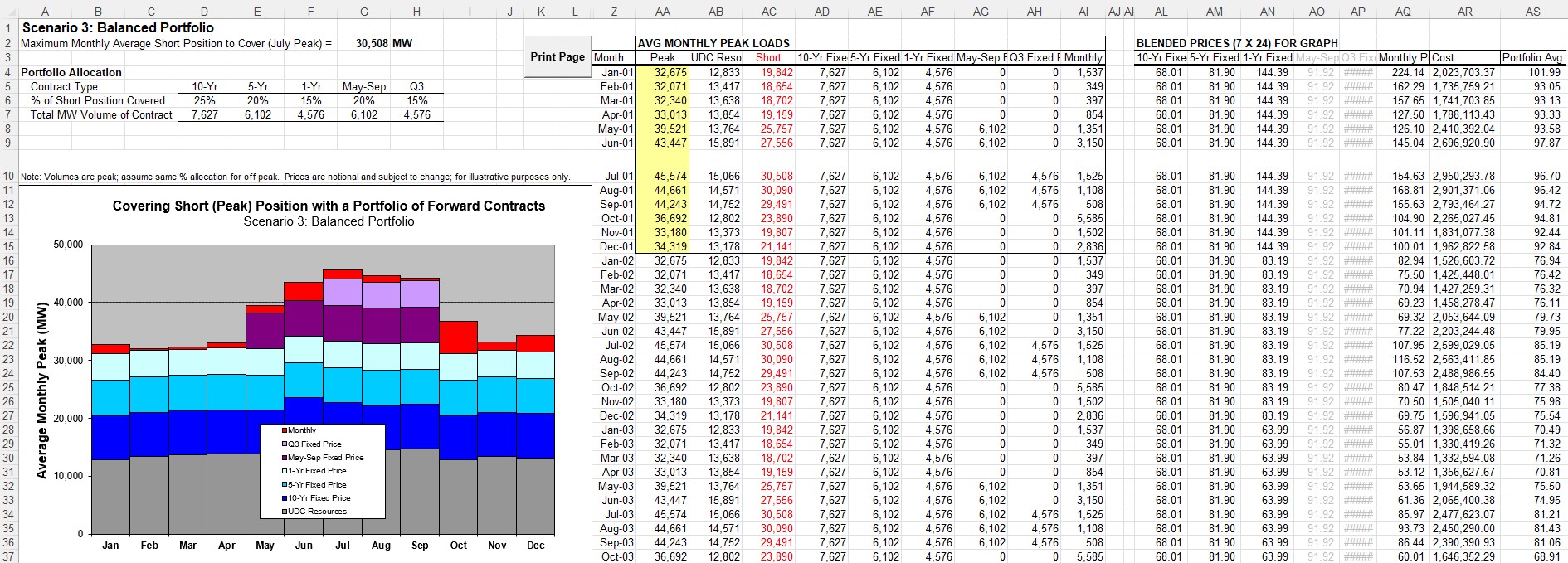}
  \captionof{figure}{This Excel sheet shows an assumption-update workflow, where a mix of forward contracts is used to cover monthly peak-load short positions. It lists the contract allocations and MW volumes, along with monthly peak loads and the resulting short MW. A table on the right computes blended prices and portfolio costs, and the stacked chart visualizes coverage by contract type over the year.}\label{fig:hy_example_1}
\end{center}

\end{document}

%% file: header.tex
\PassOptionsToPackage{numbers, compress}{natbib}

\usepackage[preprint]{neurips_2025}

\usepackage[utf8]{inputenc} 
\usepackage[T1]{fontenc}    
\usepackage{hyperref}       
\usepackage{url}            
\usepackage{booktabs}       
\usepackage{amsfonts}       
\usepackage{nicefrac}       
\usepackage{microtype}      
\usepackage{graphicx}
\usepackage{amsmath}
\usepackage{cleveref}
\usepackage{multirow}
\usepackage{colortbl}
\usepackage{wrapfig}
\usepackage{longtable}
\usepackage{gradient-text}
\usepackage{amssymb}
\usepackage{pifont}

\crefname{figure}{Fig.}{Figs.}
\Crefname{figure}{Fig.}{Figs.}
\crefname{table}{Tab.}{Tabs.}
\Crefname{table}{Tab.}{Tabs.}
\crefname{appendix}{App.}{Apps.}
\Crefname{appendix}{App.}{Apps.}
\crefformat{section}{\S#2#1#3} 
\crefformat{subsection}{\S#2#1#3}
\crefformat{subsubsection}{\S#2#1#3}

\usepackage[dvipsnames]{xcolor}
\usepackage{tikz,lipsum}
\usepackage[most]{tcolorbox}
\usepackage{listings}
\usepackage{caption}
\usepackage{fancyhdr}
\usepackage{subfigure}
\usepackage{enumitem}

\usepackage{graphicx}

\usepackage[normalem]{ulem}
\useunder{\uline}{\ul}{}

\hypersetup{
 colorlinks=True,
 linkcolor=citecolor,
 citecolor=citecolor,
 urlcolor=arxiv}

\definecolor{citecolor}{HTML}{205a88}

\definecolor{cello}{HTML}{ffe6cc}

\makeatletter
\definecolor{customblue}{HTML}{a91d3a}

\definecolor{customred}{HTML}{d62728}

\definecolor{chart}{HTML}{1f77b4}

\definecolor{arxiv}{HTML}{b31a1a}

\definecolor{sectioncolor}{HTML}{A91D3A}
\makeatother

\definecolor{OliveGreen}{rgb}{0.33, 0.42, 0.18}

\newtcolorbox{example2}[1][]{
  colback=arxiv!10!white,
  colframe=arxiv,
  floatplacement=floating,
  title=\centering #1
}

\newtcolorbox{example}[1][]{
  colback=chart!5!white,
  colframe=chart,
  floatplacement=floating,
  title=\centering #1
}

\newtcolorbox{wronganswer}[1][]{
    enhanced,
    breakable,
    colframe=arxiv,
    colback=arxiv!10!white,
    sharp corners,
    boxsep=0pt,
    left=5pt,
    right=5pt,
    top=6pt,
    bottom=6pt,
    boxrule=0pt,
    leftrule=4pt,
    #1
}

\newtcolorbox{correctanswer}[1][]{
    enhanced,
    breakable,
    colframe=OliveGreen,
    colback=OliveGreen!10!white,
    sharp corners,
    boxsep=0pt,
    left=5pt,
    right=5pt,
    top=6pt,
    bottom=6pt,
    boxrule=0pt,
    leftrule=4pt,
    #1
}

\addtocontents{toc}{\protect\hypertarget{toc}{}}

\definecolor{Gray}{gray}{0.95}

\newcolumntype{a}{>{\columncolor{Gray}}c}